\pdfoutput=1

\documentclass{article}

\PassOptionsToPackage{numbers, compress}{natbib}

\usepackage[preprint]{neurips_2026}

\usepackage[utf8]{inputenc}
\usepackage[T1]{fontenc}
\usepackage{hyperref}
\usepackage{xurl}
\usepackage{booktabs}
\usepackage{amsfonts}
\usepackage{nicefrac}
\usepackage{microtype}
\usepackage{amsmath}
\usepackage{enumitem}
\usepackage{graphicx}
\usepackage{wrapfig}
\usepackage[capitalize,noabbrev]{cleveref}
\usepackage{algorithm}
\usepackage{algpseudocode}
\usepackage{subcaption}
\usepackage{amssymb}
\usepackage{pifont}
\newcommand{\cmark}{\ding{51}}
\newcommand{\xmark}{\ding{55}}
\usepackage[table]{xcolor}
\usepackage{tabularx}
\usepackage[most]{tcolorbox}
\usepackage{adjustbox}
\usepackage{fancyvrb}
\usepackage{fvextra}
\usepackage{array}
\usepackage{multirow}
\usepackage{float}
\usepackage{makecell}
\usepackage{fontawesome5}
\usepackage{etoolbox}
\usepackage{needspace}

\newcommand{\oracle}{\textit{Oracle}}
\hypersetup{hidelinks}

\usepackage{times}

\newtcolorbox{samplecard}[1]{%
  enhanced, breakable,
  colback=gray!3, colframe=black!45,
  coltitle=black, colbacktitle=gray!15,
  fonttitle=\bfseries\small, title={#1},
  fontupper=\small,
  boxrule=0.4pt, arc=2pt,
  left=6pt, right=6pt, top=4pt, bottom=4pt,
  before skip=4pt, after skip=4pt,
}
\newcommand{\dturn}[2]{\par\smallskip\noindent\hangindent=1.4em\hangafter=1%
  \textbf{#1}\hspace{0.45em}\ignorespaces#2\par}
\newcommand{\delide}[1]{\par\smallskip\centerline{\footnotesize\itshape[#1]}\par\smallskip}
\newcommand{\dsep}{\par\smallskip\noindent\rule{\linewidth}{0.4pt}\par\smallskip}

\newtcbox{\linkbadge}{on line, nobeforeafter, arc=6pt, boxrule=0.5pt, boxsep=0pt, left=7pt, right=7pt, top=4pt, bottom=4pt, colback=black!3, colframe=black!35}
\newcommand{\hflogo}{\raisebox{-0.2em}{\includegraphics[height=1em]{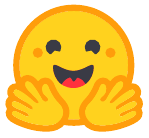}}}

\newtcolorbox{promptbox}{%
  enhanced, breakable,
  colback=gray!3, colframe=black!40,
  boxrule=0.35pt, arc=1.5pt,
  left=2pt, right=2pt, top=3pt, bottom=3pt,
  before skip=3pt, after skip=4pt,
}

\usepackage{xspace}
\newcommand{\oursys}{\textnormal{Mind2Dialogue}\xspace}
\newcommand{\oursim}{\textsc{M2D-Sim}\xspace}
\newcommand{\ourcorpus}{\textsc{M2D-Corpus}\xspace}
\newcommand{\ourmodel}{\textsc{M2D-Chat}\xspace}

\title{\textnormal{Mind2Dialogue}: Training Human-Aware Language Models by {Simulating User Mental States}}

\newcommand{\affmark}[1]{\textsuperscript{\normalfont #1}}
\author{%
\textbf{Zixuan Wang}\affmark{1$\ast$}\enspace
\textbf{Yufan Zhou}\affmark{2}\thanks{Equal contribution.\quad
\textsuperscript{\ddag}Corresponding author.}\enspace
\textbf{Jinzhou Tang}\affmark{1}\footnotemark[1]\enspace
\textbf{Xinle Yu}\affmark{1}\enspace
\textbf{Chengjun Wu}\affmark{1} \\
\textbf{Lyumanshan Ye}\affmark{1}\enspace
\textbf{Zhaoxiang Feng}\affmark{1}\enspace
\textbf{Letian Peng}\affmark{1}\enspace
\textbf{Adyasha Patra}\affmark{1} \\
\textbf{Fan Bai}\affmark{5}\enspace
\textbf{Enze Ma}\affmark{3}\enspace
\textbf{Zhengding Hu}\affmark{1}\enspace
\textbf{Jianyang Gu}\affmark{4}\enspace
\textbf{Zhao Wang}\affmark{1} \\
\textbf{Yufei Ding}\affmark{1}\enspace
\textbf{Jingbo Shang}\affmark{1}\enspace
\textbf{Tianmin Shu}\affmark{5}\enspace
\textbf{Zhiting Hu}\affmark{1}\enspace
\textbf{Zhen Wang}\affmark{1\ddag} \\[2pt]
\affmark{1}UC San Diego\enspace
\affmark{2}KU Leuven\enspace
\affmark{3}University of Illinois Chicago \\
\affmark{4}The Ohio State University\enspace
\affmark{5}Johns Hopkins University \\[2pt]
\texttt{ziw178@ucsd.edu}, \texttt{zhenwang.work@gmail.com}
}

\begin{document}
\raggedbottom

\maketitle
\setcounter{footnote}{0}

\vspace{-30pt}
\begin{center}
  \small
  \begingroup
  \hypersetup{pdfborder={0 0 0}}
  \href{https://wannabeyourfriend.github.io/mind2dialogue/}{\linkbadge{\faGlobe~Project}}\hspace{0.6em}
  \href{https://github.com/wannabeyourfriend/mind2dialogue}{\linkbadge{\faGithub~Code}}\hspace{0.6em}
  \href{https://huggingface.co/datasets/wannabeyourfriend-hf/mind2dialogue}{\linkbadge{\hflogo~Dataset}}
  \endgroup
\end{center}

\begin{abstract}

  \begingroup\par\begingroup
{
As language models become more capable, {long-term collaboration} in learning, reasoning, and decision-making calls for a deeper understanding of the people they serve.
Yet training such \emph{human-aware} language models faces a fundamental supervision gap because current datasets for LLM assistant training contain {few if any} well-informed responses explicitly grounded in users' unspoken beliefs and goals.
Scaling such supervision {is inherently constrained}, as users' underlying states are not directly observable.
We thus propose the \oursys{} framework to mitigate this gap by simulating users' mental states and turning them into privileged supervision for human-aware language model training.
Specifically, we first propose a psychology-guided simulator that preserves personal characteristics while updating mental states through interaction to generate coherent conversations.
The key idea is to enforce a shared evolving mental state that drives user behavior and guides an \oracle{} assistant's responses.
Our privileged distillation then trains models on the \oracle{}'s well-informed responses to assist users without direct access to their mental states at deployment.
Moreover, we propose to evaluate human-aware learning through the combination of both personalization and theory of mind, examining how models understand people and act on that understanding.
Training on the full \oursys{} corpus improves every reported personalization metric over the corresponding Qwen, Llama, and OLMo instruction-tuned baselines, including gains of 26.6 to 40.9 percentage points in preference-following generation.
The gains extend to belief and action reasoning on Qwen and Llama, demonstrating benefits beyond personalized assistance.
Looking forward, \oursys{} makes user simulation a foundation for genuine AI collaborators that understand beliefs and intentions behind people's words and support their long-term goals across education, work, and everyday life.
}
\looseness=-1
\par\endgroup\endgroup

\end{abstract}

\begingroup
\setlength{\parskip}{4.5pt plus 0.5pt minus 0.5pt}
\setlength{\textfloatsep}{12pt plus 2pt minus 2pt}
\setlength{\floatsep}{10pt plus 2pt minus 2pt}
\setlength{\intextsep}{10pt plus 2pt minus 2pt}
\begin{figure}[t]
  \centering
  \includegraphics[width=1.0\columnwidth]{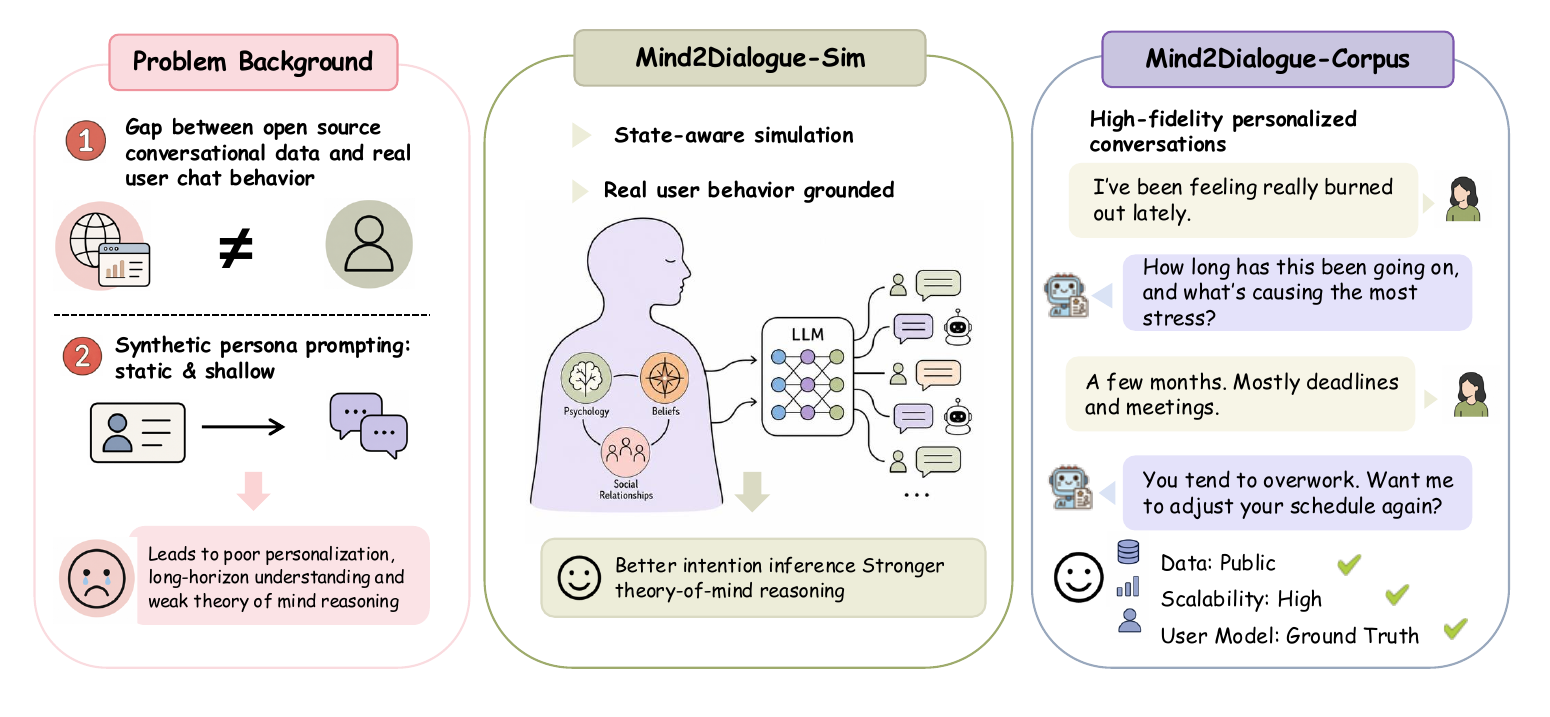}
  \caption{{\textbf{Training human-aware language models with \oursys{}.}} {\emph{Left:} existing dialogue data and static persona prompting provide limited supervision for understanding users. \emph{Middle:} \oursim{} generates user behavior and informed \oracle{} responses from shared evolving states. \emph{Right:} \ourcorpus{} supports scalable assistant training. ``Ground Truth'' denotes simulator-defined states withheld from students.}}
  \label{fig:overview}
\end{figure}

\section{Introduction}
\label{sec:introduction}

{Language models should help people learn, reason, and make decisions by accounting for the beliefs, goals, and circumstances that shape their actions.
Understanding {other people's minds} is a core component of human intelligence and motivates \emph{human-aware} capabilities in the next generation of language models~\citep{collins2024people}.
In real-world deployment, useful assistance must reflect the knowledge, priorities, and constraints of the person using the model~\citep{clark1991grounding,collins2024people}.
Sustained human-aware assistance {thus} requires adapting to \emph{evolving user states} as beliefs, goals, and circumstances change{~\citep{collins2024people,ma2026memprobe}}.}

{However, training} human-aware language models faces a {fundamental} supervision gap.
{The challenge is to obtain, at scale, training responses grounded in a deep understanding of users' unspoken beliefs and goals.}
{On the one hand,} conversations between people who know one another well provide natural examples of assistance informed by {such} mutual understanding.
Close friends, family members, and longtime collaborators{, for example,} draw on \emph{shared experience} to recognize the goals behind a request and respond with knowledge of the person's circumstances~\citep{clark1991grounding}.
{Yet collecting these private exchanges and documenting their shared background requires consent and substantial annotation effort, limiting collection at scale~\citep{ma2026synthetic,jandaghi2024faithful}.}
{On the other hand, public dialogue corpora offer scale~\citep{zhao2024wildchat}, but the user's \emph{evolving states} are not directly observable~\citep{goodman2016pragmatic}.
More surface-form dialogue alone therefore does not {teach assistants to infer and act on users' unspoken beliefs and goals.}}

Synthetic data has emerged as a promising solution to this supervision gap~\citep{ge2024scaling,ma2026synthetic}.
{Persona-conditioned generation uses descriptions of users' backgrounds and preferences to diversify synthetic conversations~\citep{ge2024scaling,jandaghi2024faithful,wang2025deeppersona}.
Profile-based generation supplies a consistent identity, but a static description leaves changes in the user's beliefs, goals, and emotions implicit in the generated exchange.
{A more recent line of work builds LLM user simulators that pursue goals and interact with off-the-shelf assistants to generate multi-turn and multi-session dialogues at scale~\citep{laban2025flipping,duan2026lifesim}.}
State modeling and simulated feedback further improve user fidelity and assistant adaptation~\citep{wu2026humanlm,kim2026propersim,zhao2025dream,luo2026userstate}.
Yet realistic user simulation does not ensure that assistants understand their users.
Evaluations with UserLM and LifeSim document failures to interpret and act on users' implicit intentions~\citep{laban2025flipping,duan2026lifesim}.
{For assistant training, realistic user simulation must also produce well-informed response targets.
When the teacher infers an unspoken state from dialogue, errors in that inference can enter the response targets.
This dependence motivates generating responses with direct knowledge of the state that shapes the user's behavior.}}
\looseness=-1

{In this paper, we propose \oursys{} to mitigate this gap by using simulated mental states to inform assistant supervision (\cref{fig:overview}).}
To construct demonstrations informed by the user's state, we introduce an {\oracle{}} assistant with direct access to that state during generation.
The key idea is \emph{shared-state user simulation}, which uses one evolving state to generate user behavior and guide the {\oracle{}}'s responses.
The {\oracle{}} can thus demonstrate how to assist a user whose beliefs, goals, and emotions are only partially expressed in the dialogue.
{The teacher can then base its response on the state that generates the interaction, without having to reconstruct that state from the dialogue.}

Scaling {this privileged supervision} requires diversity across users and coherence within each interaction.
We therefore build \oursim{} {as a psychology-guided simulator} with scenarios that give users reasons to seek assistance, state updates that track their changing circumstances, and a controller that varies their conversational behavior (\cref{fig:framework}).
The resulting \ourcorpus{} combines multi-turn {\oracle{}} dialogues with question-answer examples derived from the same interactions, without requiring human annotation for each generated dialogue.
{To transfer the {\oracle{}}'s decisions to a deployable model, we use privileged distillation to train \ourmodel{} on these responses through supervised fine-tuning~\citep{lopez-paz2016unifying}.}
The student learns from the visible inputs and target responses, with the evolving state withheld during both training and inference.
This information asymmetry allows mental states to guide what the model learns without requiring those states as inputs when the model assists a user.

{Moreover,} evaluating human-aware language models through long-term interaction with real users is difficult to scale~\citep{li2026horizonbench}.
{We bring together two seemingly distinct domains, personalization and theory of mind, to examine how models understand people and use that understanding in assistance.}
{Personalization tests whether models act on users' preferences and circumstances; theory of mind tests whether learning from simulated interaction transfers to reasoning about beliefs and actions.}
{Both domains use independently constructed benchmarks whose content is excluded from our training corpus.}
Training on \ourcorpus{} improves every measured personalization metric on PersonaMem-v1, PersonaMem-v2, and PrefEval~\citep{jiang2025know,jiang2025personamem,zhao2025prefeval} across Qwen2.5-7B, Llama-3.1-8B, and OLMo-3-7B (\cref{tab:sft_sample_scaling_unified}).
Qwen2.5-7B gains 33.4 percentage points on PrefEval generation and 10.0 points on PersonaMem-v2 multiple-choice accuracy over its base model (\cref{tab:main_personalization}).
On ToMi and BigToM~\citep{le2019revisiting,gandhi2023bigtom}, Qwen and Llama improve across all three tasks, including a 13.0-point gain for Qwen on BigToM forward-belief accuracy (\cref{tab:main_tom,tab:sft_sample_scaling_unified}).
OLMo improves on ToMi and declines on both BigToM tasks, showing that the benefits for mental-state reasoning vary across models.
{\oursys{} makes user simulation a practical route toward human-aware collaboration by turning knowledge of the person behind a request into training supervision, supporting the broader pursuit of \emph{personal AGI} in service of individual goals~\citep{altman2026plan}.}

\section{Related Work}
\label{sec:related_work}

\noindent\textbf{{Human-AI collaboration and user modeling.}}
{Research on human-AI collaboration examines how language models can work with people whose knowledge, intentions, and need for control shape the task~\citep{collins2024people,shao2026collaborative,mo2023roll}.
OpenAI's \emph{Personal AGI} agenda similarly envisions broadly capable AI that people can direct toward their own objectives~\citep{altman2026plan}.
Cooperative inverse reinforcement learning formalizes assistance under uncertainty about human preferences, making communication part of cooperative decision-making~\citep{hadfieldmenell2016cooperative}.
For language models, CollabLLM uses rewards over multiple turns to train assistants to elicit user intent and advance the user's goal~\citep{wu2025collabllm}.
Proactive Agent learns to propose assistance from user activity and environmental context before an explicit request~\citep{lu2025proactive}.
Co-Gym complements these approaches with shared workspaces for evaluating communication, coordinated action, and user control~\citep{shao2026collaborative}.
For sustained assistance, LongMemEval tests memory across sessions, while HorizonBench tests whether models track preferences as life events change user states~\citep{wu2024longmemeval,li2026horizonbench}.}

{Personalization addresses how assistance should reflect the individual within this broader collaboration problem.}
{Existing methods} augment a fixed model with external memory and retrieval~\citep{chhikara2025mem0,zhong2024memorybank,li2024ldagent} or adapt model parameters using user-specific data~\citep{salemi2024lamp,magister2024plum}.
{PersonaMem-v2 uses preference supervision for reinforcement fine-tuning and agentic memory learning~\citep{jiang2025personamem}.
DreamCUB learns a dialogue world model that predicts utterances and user beliefs for model-based reinforcement learning~\citep{zhao2025dream}; PUMA maintains beliefs over partially observed user states and plans using predicted state transitions~\citep{luo2026userstate}.
\oursys{} constructs assistant supervision by giving the teacher direct access to the simulated state that generates user behavior.}

\noindent\textbf{{Synthetic dialogue and user simulation.}}
Synthetic dialogue research {spans user-query generation}~\citep{chen2022bootstrapping}, fixed persona-conditioned generation~\citep{zhang-etal-2018-personalizing,ge2024scaling}, {and} multi-turn and multi-session LLM user simulators~\citep{davidson2023user,sekulic2024reliable,laban2025flipping,jiang2025personamem}.
{Persona Hub expands profile diversity, while Synthetic-Persona-Chat improves persona consistency through generation and critique~\citep{ge2024scaling,jandaghi2024faithful}.
Generative Agents and LifeSim extend simulation to behavior shaped by memory and changing circumstances~\citep{park2023generative,duan2026lifesim}.
UserLM learns intent-conditioned user behavior from human conversations~\citep{laban2025flipping}, and HumanLM aligns generated mental states and responses with real users through reinforcement learning~\citep{wu2026humanlm}.
HumanLM trains the simulator; \oursys{} uses simulation to train the assistant.}

{Simulated interaction also supports assistant learning through feedback and rewards.
ProPerSim adapts proactive recommendations using simulated user ratings~\citep{kim2026propersim}; PersonaGym supports personalized prompt optimization through profile inference and outcome feedback~\citep{ma2026synthetic}.
UserRL trains interactive agents with simulated users and studies turn-level rewards and trajectory scoring~\citep{qian2025userrl}.}
We focus on \emph{who observes the simulator-defined user state when assistant responses are generated}.
{The {\oracle{}} observes the state that drives user behavior before generating the response target.}

\noindent\textbf{{Social intelligence and mental-state reasoning.}}
{Social intelligence research examines both inferring other agents' states and using that understanding to act.
Machine Theory of Mind learns to predict agents' behavior and mental states from observed trajectories~\citep{rabinowitz2018machine}; SOTOPIA-$\pi$ trains language agents through behavior cloning and self-reinforcement on interactions selected by social-goal ratings~\citep{wang2024sotopia}.
SimpleToM shows that accurate state attribution can coexist with errors in predicting or judging behavior~\citep{gu2026simpletom}, motivating our {complementary evaluation of assistance and reasoning.}}

{{Mental-state annotations and state-informed assistant demonstrations provide different forms of supervision.}
ToMATO combines personas with turn-level first- and second-order thoughts, keeping each speaker's thoughts hidden from its partner.
Those thoughts supply mental-state QA labels for evaluation and for fine-tuning on separately generated conversations{~\citep{shinoda2025tomato}}.
Our {\oracle{}} observes the evolving state that produces the user's behavior and uses it to generate response demonstrations for a student with that state withheld.
}

\noindent\textbf{{Learning with privileged information.}}
{Learning with privileged information and generalized distillation allow a teacher to use information unavailable to the student~\citep{vapnik2009new,hinton2015distilling,lopez-paz2016unifying}.}
Whereas contemporaneous work uses joint or on-policy objectives to transfer from privileged policies~\citep{penaloza2026privileged}, we rely on a fixed {\oracle{}} and standard supervised fine-tuning.
{The privileged information in \oursys{} is the state that generates the interaction itself.
{Sharing this state with the {\oracle{}} gives the teacher direct access to the beliefs and goals that generate user behavior, while the student receives only observable inputs and target responses.}
Our contribution is the construction of this supervision through an integrated simulator, corpus, and assistant-training pipeline.}

\Needspace{5\baselineskip}
\section{{The \oursys{} Framework}}
\label{sec:method}

{\oursys{} is a framework for training human-aware language models with supervision informed by simulated user states.}
{The rollout engine in \cref{fig:framework} generates multi-turn training data through interaction between a user simulator and an \oracle{} assistant.}
During data generation, both components have access to a persona $p$ and a shared structured state $s_t$.
{\oursim{} is the simulator that generates these interactions.
\ourcorpus{} contains these dialogues and derived question-answer examples.
\ourmodel{} denotes the student language models trained on this corpus with the evolving state withheld.}
{The teacher and student views distinguish state access during generation from the observable inputs used for training and deployment.}
\looseness=-1

\begin{figure}[!t]
  \centering
  \includegraphics[width=\linewidth]{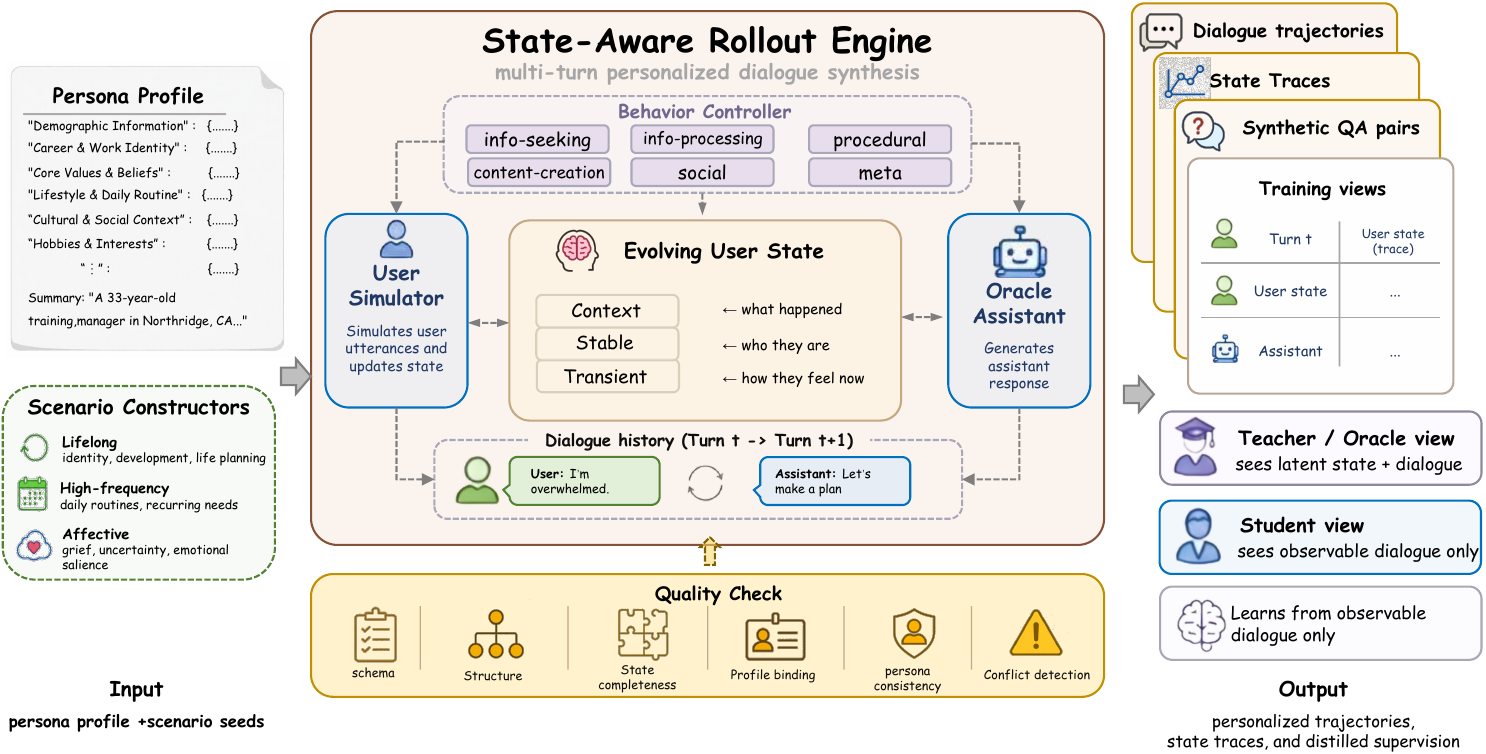}
  \caption{{\textbf{Psychology-guided simulation and privileged supervision.}} {\emph{Left:} persona profiles and scenario seeds initialize the interaction. \emph{Center:} the rollout engine uses one evolving state to generate user messages and \oracle{} responses, with a controller varying user behavior. \emph{Below:} quality checks filter trajectories. \emph{Right:} retained dialogues and derived QA supervise students, with state traces withheld from their inputs.}}
  \label{fig:framework}
\end{figure}

\Needspace{6\baselineskip}
\subsection{{Framework Overview and Supervision Design}}
\label{sec:problem_setup}

\noindent\textbf{{Problem formulation.}}
{Human-aware training must teach assistants to respond to users whose beliefs, goals, and circumstances are only partly expressed in conversation.}
Let $p$ denote the persona used to generate a dialogue, $H_{<t}$ the history of the dialogue before the $t$-th user message and $m_t$ the user message at that turn.
We define
\[
  H_t = (H_{<t}, m_t)
\]
as the dialogue history through the user message.
{Let $s_t$ represent the user's evolving state and $x_t$ the student input, which includes $H_t$ and excludes $s_t$.
{We aim to learn} an assistant policy $\pi_\phi(a_t\mid x_t)$ that uses the available evidence about the user's state to guide its responses.
{The student has no direct state access during either training or deployment.}}

{The supervision gap arises when the response target depends on a user state that the input does not fully specify.
A teacher restricted to $x_t$ must infer the missing state before choosing a response, making supervision depend on the teacher's existing understanding of users.
We seek demonstrations whose targets are generated with knowledge of $s_t$, while retaining $x_t$ as the student's input.}

\noindent\textbf{{Shared-state {\oracle{}} supervision.}}
{\oursys{} addresses this problem by generating the user state and the dialogue together.
The key idea is to share that state with the assistant before it produces a training target.
We use \emph{Oracle} to denote direct access to the simulated state during response generation.
A common persona alone does not specify how beliefs and goals change within the interaction; sharing $s_t$ gives both policies the same evolving account of those changes.}
{At each turn, $\pi_{\mathrm{state}}$ updates the user's state, $\pi_{\mathrm{user}}$ generates a message, and $\pi_{\mathrm{oracle}}$ produces the response target.
These policies denote distinct generation steps that can share a language-model backbone.}
\begin{align*}
  {s_t} &{\sim \pi_{\mathrm{state}}(\,\cdot\mid p,s_{t-1},H_{<t})}, \\
  m_t &\sim \pi_{\mathrm{user}}(\,\cdot\mid p,s_t,H_{<t}), \\
  a_t &\sim \pi_{\mathrm{oracle}}(\,\cdot\mid p,s_t,H_t).
\end{align*}
{The shared $s_t$ connects the cause of the simulated behavior with the information used to produce its response.
The next state update observes $H_{<t+1}=(H_t,a_t)$, including the assistant response.
Subsequent turns can therefore reflect changes induced by the assistant's response.}

From $N$ {retained} dialogues, we construct the dialogue training set
\[
  \ourcorpus_{\mathrm{dialogue}}
  =\left\{\bigl({x_t^{(i)}},a_t^{(i)}\bigr)
  \;\middle|\;1\leq i\leq N,\;1\leq t\leq T_i\right\},
\]
where $T_i$ is the number of assistant-response targets in dialogue $i$.
{Each response target is thus paired with the evidence the student will observe, while its construction also uses the underlying simulated state.
\Cref{sec:ours,sec:corpus_construction} describe how we control these trajectories and retain coherent demonstrations; \cref{sec:student_training} formalizes learning with the state withheld.}

\subsection{{Psychology-Guided Stateful User Simulation}}
\label{sec:ours}

{We ground the design of \oursim{} in psychological accounts of how personal characteristics and situational demands jointly shape behavior~\citep{funder2006personalitytriad}.}
The cognitive-affective processing account further describes how situational features interact with goals, affect, expectations, and related internal variables to produce context-dependent behavior~\citep{mischel1995cognitiveaffective}.
{These accounts motivate a simulator that preserves personal characteristics while allowing internal states and behavior to change with the interaction.}
In our implementation, the scenario supplies the immediate context for the interaction, $s_t$ records simulator-defined variables that carry across turns, and dynamic behavior-mode prompting controls how the user acts at each turn.
{\oursim{} makes these controls explicit to preserve continuity while varying the situations and behaviors represented in the training data.}

\noindent\textbf{{Persona-grounded scenario construction.}}
The scenario specifies why the user starts the interaction and provides a setting in which persona-specific information can affect how the assistant should respond.
Each dialogue begins with a scenario derived from the persona $p$.
We construct scenarios in three categories: \textsc{Lifelong}, covering identity and long-term personal development; \textsc{High-Frequency}, covering recurring everyday needs; and \textsc{Affective}, covering emotionally significant situations such as grief or uncertainty.
We filter candidate scenarios using three criteria: level of abstraction, embedding-based distinctness from existing scenarios, and semantic consistency with the persona.
For the distinctness check, we reject a candidate if its cosine similarity to any existing scenario exceeds $\theta_{\mathrm{sim}}$.
Accepted scenarios are cached for each persona so that subsequent runs can reuse the same scenario set.

\Needspace{6\baselineskip}
\noindent\textbf{{Evolving user-state simulation.}}
{An assistant's response can change what a user believes or needs, so the simulator must carry those changes into subsequent turns.}
\oursim{} represents the state shared by the generation policies as {a structured record}
\[
  s_t = \bigl(c_t,\; z_t^{\mathrm{stable}},\; z_t^{\mathrm{transient}}\bigr),
\]
{where $c_t$ tracks the turn index, unresolved goals, trust history, and a summary of the interaction.}
The component $z_t^{\mathrm{stable}}$ stores information that changes slowly, such as user values, background constraints, and the user's position toward the assistant.
The component $z_t^{\mathrm{transient}}$ stores short-lived information, such as mood, current concerns, and judgments made during the turn.
These fields are simulator-defined control variables, not measurements of a real user's mental state.

\noindent\textbf{{Turn-level behavior control.}}
{Users can express a goal through questions, requests, or reactions to an assistant, so varied interactions also require control over conversational behavior.}
{Our behavior controller builds on the Taxonomy of User Needs and Actions (TUNA)}~\citep{shelby2025taxonomyuserneedsactions}.
{The controller selects among $14$ TUNA-derived modes and two fallback modes. The six TUNA families cover} information seeking, information processing, procedural guidance, content creation, social interaction, and meta-conversation.
{The controller adds mode-specific instructions to the user-simulator prompt, with less behavioral guidance at turn~1 and more at later turns or when the user delegates more to the assistant.}
{The selection procedure} also encourages coverage across the six families.
The full taxonomy and mode descriptions appear in~\cref{app:behavior_modes}.

\Needspace{5\baselineskip}
\subsection{{Privileged Supervision and Corpus Construction}}
\label{sec:corpus_construction}

{A simulated interaction supplies both examples of informed assistance and the context needed to ask questions about the user.}
{We retain both views to teach response generation within an interaction and the use of user information in explicit question answering.}
{The dialogue view pairs the student-visible context with the {\oracle{}} response.}
{The QA view uses the persona, saved state trajectory, and a dialogue excerpt to generate questions and answers about the simulated interaction.}
We generate multiple-choice persona-memory examples and free-form preference-following examples using the same message schema as the dialogue examples.
{The mixture also contains preference-classification questions.
Student QA inputs contain the observable context, the question, and any answer options required by the format; the structured state remains available only during generation.
\Cref{app:stats_mixture} reports the composition of the training mixture.}
\looseness=-1

{Training requires coherent trajectories that preserve the simulated user's context.}
We apply four programmatic checks to generated conversations: schema validity, structural sanity (turn count, role alternation, and token bounds), state-trajectory completeness, and profile binding.
Two LLM-based checks additionally score persona consistency and flag conflicts with fixed persona attributes, using a judge model distinct from the simulator and the {\oracle{}}.
\Cref{app:qc} describes the checks and reports a human audit of the filtered corpus; the two judge rubrics are {provided} in~\cref{app:qc_prompts}.
{\Cref{sec:appendix-qualitative} illustrates how response targets reflect the simulated user's background and priorities.}
\looseness=-1

\subsection{{Privileged-State Supervised Distillation}}
\label{sec:student_training}

{Human-aware assistance requires learning response decisions under partial observability of the user's state.
We use the {\oracle{}}'s state access to construct targets while keeping the student's information constraints identical at training and inference.
The setup follows generalized distillation~\citep{lopez-paz2016unifying}, with privileged information supplied through response targets.
{The privilege lies in the state information used to construct the response target, even when the {\oracle{}} and user simulator share a backbone.}}

\begingroup\par\begingroup
The student learns the {\oracle{}}'s response behavior through the evidence available in its input.
For dialogue demonstrations, let $q(x,s,y)$ denote the joint distribution of student inputs, simulated states, and {\oracle{}} responses induced by simulation and quality filtering.
{Conditioning on $x$ yields}
\begin{equation}
  \bar q(y\mid x)
  = \mathbb{E}_{s\sim q(\cdot\mid x)}\!\left[q(y\mid x,s)\right].
  \label{eq:student_target}
\end{equation}
The conditional $q(s\mid x)$ accounts for states consistent with the visible context, and $q(y\mid x,s)$ captures the corresponding state-informed targets.
These distributions describe the generated data and are not separately estimated during training.
\par\endgroup\endgroup

We fine-tune a student policy $\pi_\phi$ on $\ourcorpus$.
Each training example consists of a student-visible input sequence $x$ and a target assistant response $y$.
\par\Needspace{7\baselineskip}
{We minimize the standard autoregressive cross-entropy loss over the target response tokens}
\begin{equation}
  \mathcal{L}_{\mathrm{SFT}}(\phi)
  =-\mathbb{E}_{(x,y)\sim\ourcorpus}
  \left[\sum_{j=1}^{|y|}\log\pi_\phi(y_j\mid x,y_{<j})\right].
  \label{eq:sft}
\end{equation}
Only tokens in the target response $y$ contribute to the loss.

{For dialogue examples, minimizing the population loss is equivalent to minimizing the expected Kullback-Leibler divergence
$\mathbb{E}_{x\sim q}D_{\mathrm{KL}}\!\left(\bar q(\cdot\mid x)\,\|\,\pi_\phi(\cdot\mid x)\right)$.
An unrestricted policy therefore has the optimum $\pi^*(\cdot\mid x)=\bar q(\cdot\mid x)$ on the support of $q$.}
{
{The {\oracle{}}'s state access determines the supervision, while the student's visible context determines the response distribution it can learn.
{The objective supervises responses without prescribing an internal state representation.}}}

\Needspace{5\baselineskip}
\section{{Training Human-Aware Language Models}}
\label{sec:experiments}

{We develop \ourmodel{} by training language models on the supervision generated by \oursim{}. This section presents the training data and recipe, {tests whether simulations sustain personal context for supervision}, and evaluates the resulting models on personalization and mental-state reasoning.}

\subsection{{Training Data and Recipe}}
\label{sec:corpus_coverage}

{Running \oursim{} over $289$ personas yields \ourcorpus{}, including a subset of $6{,}330$ multi-turn conversations analyzed here. We characterize scenario coverage, persona specializations, and user behavior. Scenario categories are recorded during generation; the behavioral analysis classifies generated user messages by embedding-based matching to mode descriptions (\cref{app:stats_modes}).}

\begin{figure}[htbp]
  \centering
  \begin{minipage}[t]{0.55\linewidth}
    \centering
    \includegraphics[width=\linewidth]{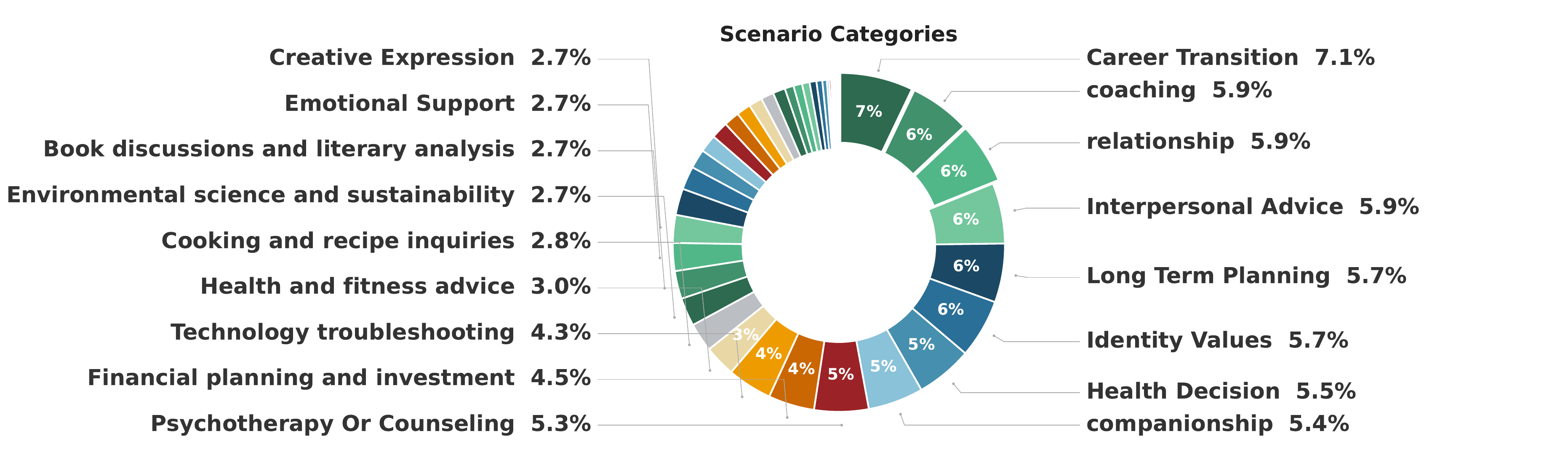}
  \end{minipage}%
  \hfill
  \begin{minipage}[t]{0.44\linewidth}
    \centering
    \includegraphics[width=\linewidth]{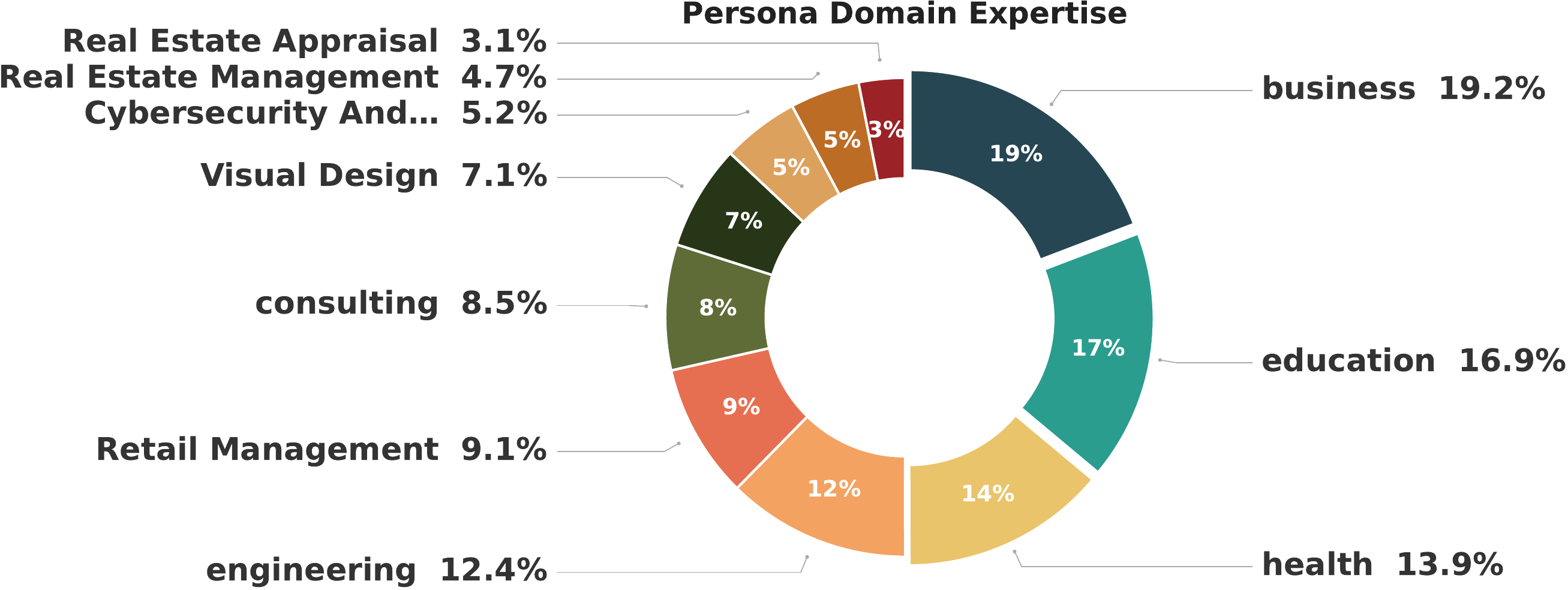}
  \end{minipage}
  \caption{{\textbf{Scenario and persona coverage.} The $6{,}330$-conversation analysis subset spans $47$ scenario categories and $10$ persona-specialization clusters. {\emph{Left:}} Scenario shares across affective, high-frequency, and lifelong conversations; labels mark categories with shares of at least $2.7\%$. {\emph{Right:}} Specializations grouped by embedding similarity. Both panels report shares of conversations in the analysis subset.}}
  \label{fig:dataset_composition}
\end{figure}

\noindent\textbf{Scenarios.} Every conversation carries a scenario category assigned by its constructor. The subset spans $47$ categories in the three families introduced above: affective ($1{,}428$ conversations), high-frequency ($2{,}020$), and lifelong ($2{,}882$). {Category coverage is long-tailed. The top} $9$, $18$ and $28$ categories represent ${\sim}52\%$, $81\%$ and $96\%$ of the data (\cref{fig:dataset_composition}, {left}).

\noindent\textbf{Personas.} Each persona declares a free-form \texttt{specialization}; the pool contains $153$ distinct values, which we cluster into $10$ groups for visualization (\cref{fig:dataset_composition}, {right}). The personas are grounded in five focal countries, and the broader rollout corpus extends to a global pool.

\begin{figure}[t]
  \centering
  \includegraphics[width=0.84\linewidth]{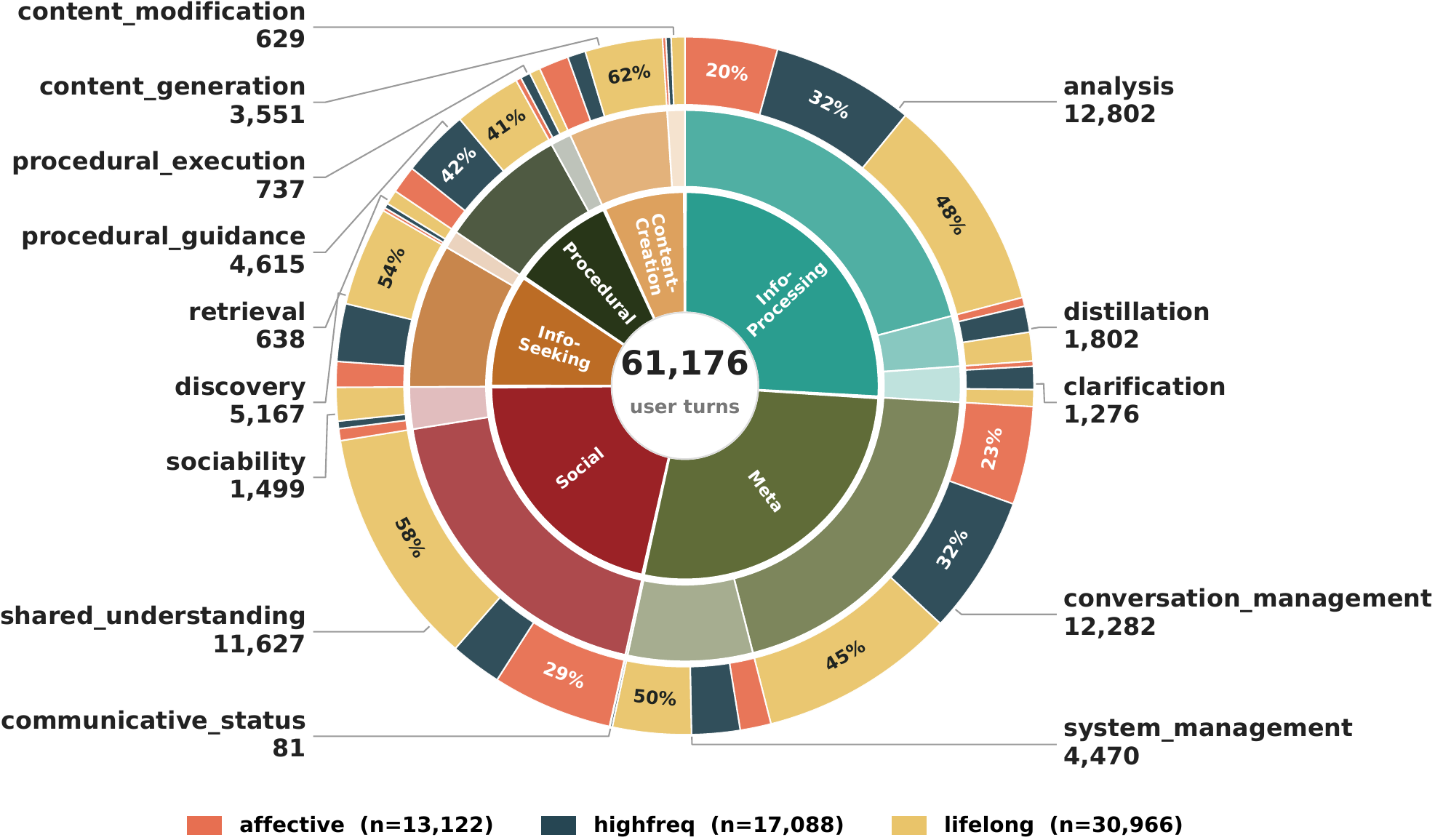}
  \caption{{\textbf{Behavioral coverage of simulated users.} The $61{,}176$ plotted turns cover $14$ modes across six behavior families. The rings show families, modes, and conversation sources from the center outward. Sector sizes encode turn counts; percentages are source shares within each mode. Mode assignments follow the embedding-based classifier in \cref{app:stats_modes}.}}
  \label{fig:mode-distribution}
\end{figure}

{The behavioral analysis measures the content of generated user turns independently of the controller's intended mode. \Cref{fig:mode-distribution} reports counts for the $14$ displayed modes across six families, with each mode decomposed by conversation source. \Cref{app:behavior_modes} includes the two fallback modes.}

{The generation pipeline supports additional personas, scenario categories, and behavioral modes without human annotation of each generated dialogue. \Cref{sec:dataset_statistics} reports detailed corpus statistics.}

\noindent\textbf{{Training mixture.}}
{The SFT mixture contains $8{,}244$ examples, comprising $3{,}312$ dialogue examples and $4{,}932$ QA examples for persona memory and preference generation or classification (\cref{tab:sft_mixture}).}

\noindent\textbf{Training.}
Our primary backbone is Qwen2.5-7B-Instruct~\citep{qwen2025qwen25technicalreport}.
To test whether data scaling depends on the backbone, we also train Llama-3.1-8B-Instruct~\citep{grattafiori2024llama} and OLMo-3-7B-Instruct~\citep{olmo2025olmo}.
{We fine-tune each open-source backbone using four fractions of the SFT mixture}: $\nicefrac{1}{8}$, $\nicefrac{1}{4}$, $\nicefrac{1}{2}$, and \textsc{All}. {We keep the training recipe fixed across model families and data scales.} Training uses LoRA~\citep{hu2022lora} with 4-bit quantization~\citep{dettmers2023qlora}, response-only masking, and AdamW~\citep{loshchilov2017decoupled} with a cosine learning-rate schedule.

\Needspace{5\baselineskip}
\subsection{{Pilot Study of Simulation Quality}}
\label{sec:preliminary}

{{We test whether simulation controls sustain personal context in user messages and assistant responses across turns.} {The \oursim{} and \textit{Vanilla} conditions} use the same backbone and persona, while Vanilla omits structured state maintenance and behavior guidance. This comparison evaluates the two components together. \Cref{app:pilot_tables} reports {detailed judge scores and a component ablation.}}

\begin{figure*}[t]
  \centering
  \includegraphics[width=\linewidth]{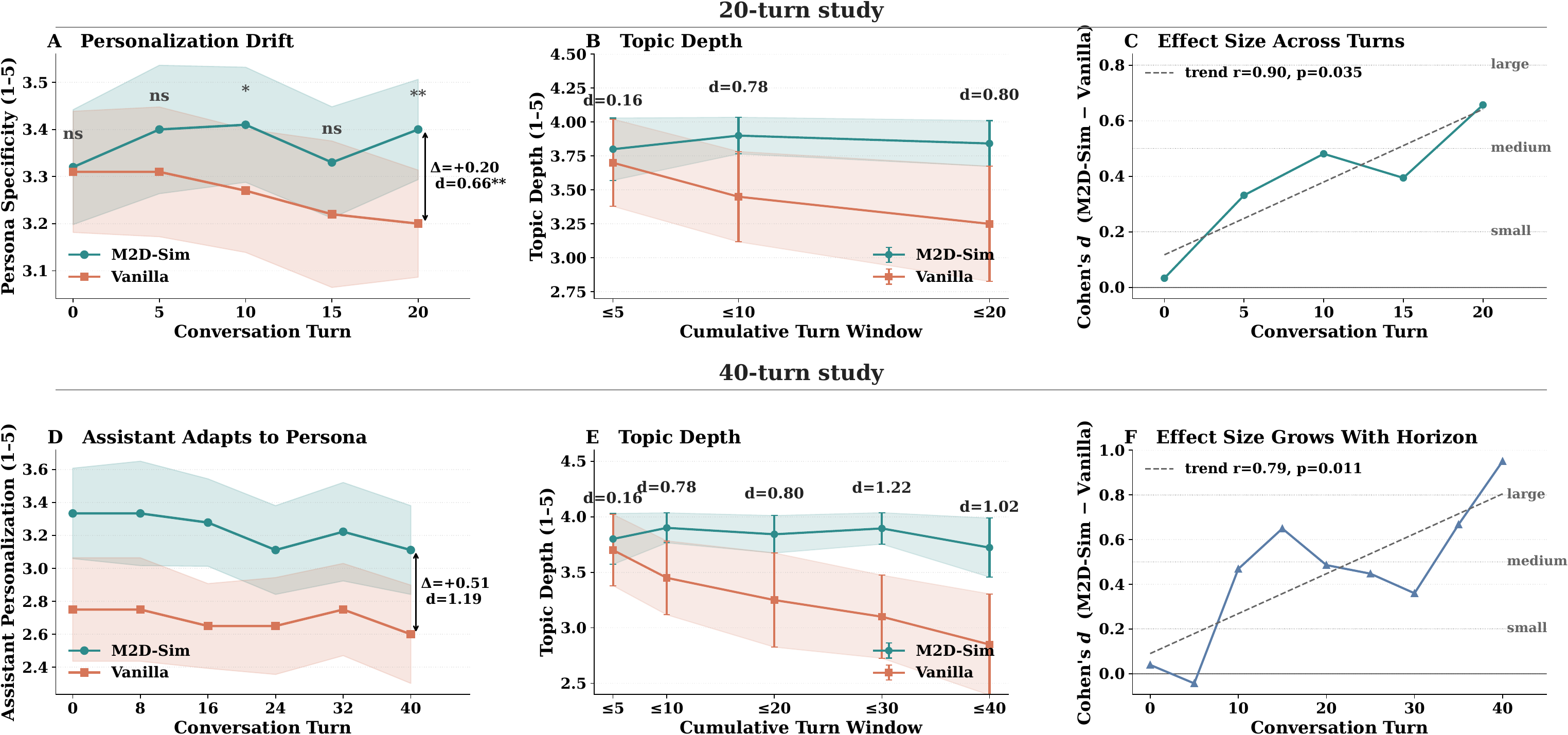}
  \caption{{\textbf{Simulation quality across conversation turns.} \oursim{} shows larger advantages over Vanilla at later turns with backbones and personas held fixed. \textbf{A} measures user persona specificity and \textbf{D} assistant personalization; \textbf{B, E} measure topic depth over cumulative windows. Judge scores range from $1$ to $5$. \textbf{C, F} report Cohen's $d$, with positive values favoring \oursim{}; $r$ gives the correlation with turn index. Rows show $20$ turns (top) and $40$ turns (bottom). Vanilla omits both state maintenance and behavior guidance.}}
  \label{fig:ablation_state_prompting}
\end{figure*}

{The pilot shows larger differences favoring \oursim{} at later stages of interaction (\cref{fig:ablation_state_prompting}). Panel~\textbf{A} marks the initial difference in user persona specificity as nonsignificant and reports $d{=}0.66$ at turn~$20$. Topic-depth differences increase from $d{=}0.16$ over the first five turns to $d{=}1.22$ over the first $30$ turns, then reach $d{=}1.02$ over all $40$ turns (\textbf{E}). Panel~\textbf{D} separately measures assistant personalization and reports a final score gap of $0.51$. The effect-size trajectories in \textbf{C} and \textbf{F} have positive correlations with turn index ($r{=}0.90$ and $r{=}0.79$), without increasing at every measured turn.}

\Needspace{5\baselineskip}
\subsection{{Evaluation Protocol}}
\label{sec:setup}

{We design the evaluation around two complementary requirements for human-aware models, applying personal context in assistance and reasoning about other agents' mental states.
Personalization benchmarks assess the use of user information in responses; Theory of Mind (ToM) benchmarks assess belief attribution and its consequences for action.
Because accurate state attribution can coexist with errors in applying that understanding~\citep{gu2026simpletom}, gains in either domain alone leave an incomplete picture.
Evaluating both domains under the same training tests the breadth of learning from simulated interaction.
A separate human audit assesses the quality of the training supervision.}

\Needspace{4\baselineskip}
\noindent\textbf{{Human evaluation of training data.}}
{A human audit evaluates whether the filtered conversations provide coherent, state-consistent demonstrations.
Annotators assessed a uniform sample of $1{,}240$ retained conversations using a binary rubric covering persona consistency, trajectory coherence, state-response consistency, and response relevance.
Of these, $1{,}216$ passed ($98.06\%$), supporting the quality of the supervision supplied to the student (\cref{app:qc}).}

\noindent\textbf{{Personalization benchmarks.}}
{The personalization suite tests whether models apply conversational context to a user's subsequent request.
{\textit{PersonaMem-v1}} tests memory and adaptation to changing profiles~\citep{jiang2025know}; {\textit{PersonaMem-v2}} emphasizes implicit preferences in task-oriented dialogue~\citep{jiang2025personamem}.
{\textit{PrefEval}} tests adherence to preferences conveyed explicitly or indirectly in earlier conversation~\citep{zhao2025prefeval}.
We report answer-selection and generation scores separately to distinguish recognizing a suitable response from producing one, using the task and backbone coverage in \cref{tab:sft_sample_scaling_unified}.}

\noindent\textbf{{Theory of mind benchmarks.}}
{The ToM suite tests whether models track another agent's knowledge and its consequences for action.
{\textit{ToMi}} evaluates first- and second-order beliefs in narratives with unequal access to information~\citep{le2019revisiting}.
{\textit{BigToM}} links percepts, beliefs, desires, and actions through a causal scenario structure~\citep{gandhi2023bigtom}.
We report its Forward Belief and Forward Action tasks to assess both state attribution and behavior prediction.
{For each task, paired true-belief and false-belief accuracy ($\mathrm{TB}\wedge\mathrm{FB}$) counts a scenario as correct only when both variants are answered correctly.
The variants differ in perceptual access, testing whether predictions reflect the agent's information.}}

\noindent\textbf{{Baselines and reference models.}}
We compare \ourmodel{} with the unmodified backbone and with task-specific methods implemented on the same backbone. The personalization baselines are PersonaVLM~\citep{nie2026personavlm}, HumanLM~\citep{wu2026humanlm}, LLMoPt~\citep{ma2026synthetic}, and Qwen2.5-7B-Instruct augmented with Mem0~\citep{chhikara2025mem0}; the ToM baselines are AutoToM~\citep{zhang2025autotom} and {ThoughtTracing}~\citep{kim2025hypothesis}. We include GPT-4o-mini~\citep{openai2024gpt4omini} and GPT-5-mini~\citep{singh2025openai} as proprietary reference points.

\noindent\textbf{{Model comparisons and metrics.}}
{The method comparison uses Qwen2.5-7B-Instruct; the scaling study compares Qwen, Llama, and OLMo with their own unmodified backbones at the training fractions in \cref{sec:corpus_coverage}.
We report scores as percentages and gains as percentage-point differences, retaining separate results for each task and backbone.
\Cref{tab:sft_sample_scaling_unified} provides the full scaling results, and \cref{tab:cross_backbone_gains} summarizes the metrics shared by all three backbones.}

\noindent\textbf{{Training data and benchmark overlap.}}
{All benchmark content is held out from corpus generation.
The PersonaMem-style QA component shares the four-option evaluation interface, while its user histories, questions, and answers come from independently simulated interactions.
We therefore interpret PersonaMem MCQ as evaluation on new content under a familiar response format.
\Cref{app:stats_mixture,app:data_provenance} detail the training mixture and the exact-overlap audit.}

\subsection{{Evaluation Results}}
\label{sec:main_results}

\noindent\textbf{Scaling \ourcorpus{} improves personalization across model families.} \oursim{} can expand the training data without human annotation {per dialogue.} We tested scalability by training three open-source backbones {using} four fractions of the same \ourcorpus{} mixture. For each {backbone}, the full mixture is best in every reported personalization metric (\cref{fig:scaling-in-domain,tab:sft_sample_scaling_unified}). PrefEval-Gen gains most ($26.6$ to $40.9$ points); the two PersonaMem metrics improve by smaller margins across all three backbones. The trajectories {differ across backbones.} Qwen records its largest PrefEval-Gen gain at the final scale, Llama realizes most of its gain by $\nicefrac{1}{4}$, and OLMo improves more evenly. The sweep supports \ourcorpus{} as scalable personalization supervision in the corpus sizes tested.

\begin{figure}[htbp]
  \centering
  \includegraphics[width=1.0\linewidth]{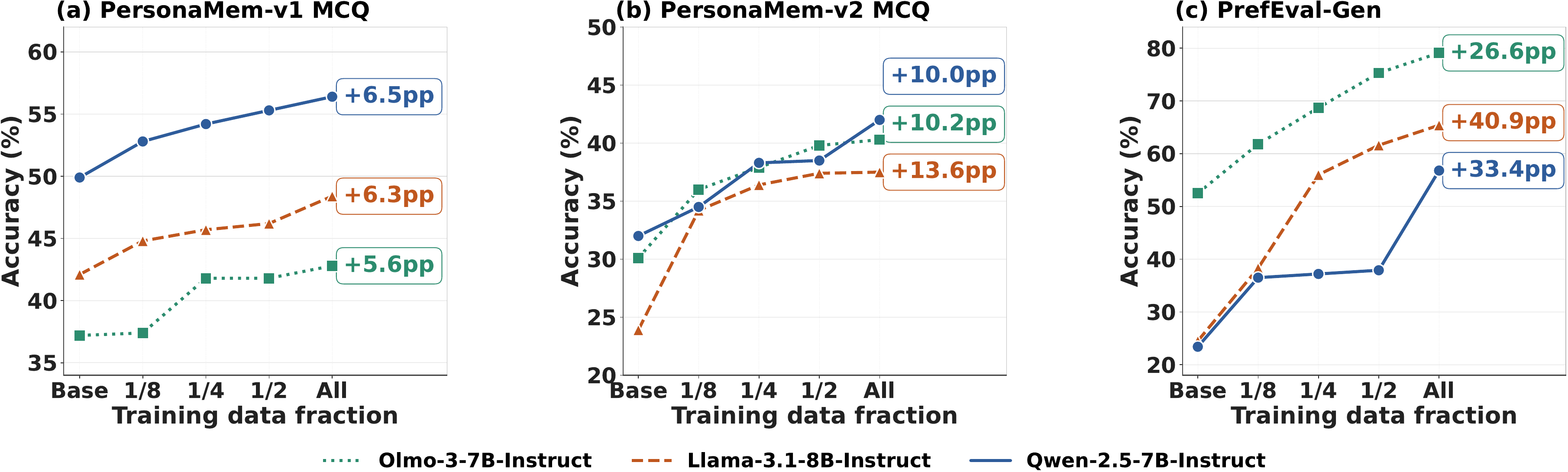}
  \caption{{\textbf{Personalization performance across training scales.} The full \ourcorpus{} mixture gives the highest accuracy on each plotted metric for all three backbones.} Accuracy is reported on \textbf{(a)} PersonaMem-v1 MCQ, \textbf{(b)} PersonaMem-v2 MCQ, and \textbf{(c)} PrefEval-Gen. Labels report the absolute gain {in percentage points} from the unmodified backbone to the full SFT mixture; \cref{tab:sft_sample_scaling_unified} gives the underlying values.}
  \label{fig:scaling-in-domain}
\end{figure}

\noindent\textbf{\ourmodel{} outperforms the evaluated personalization baselines.} On Qwen2.5-7B-Instruct, \ourmodel{} achieves the best non-proprietary score in all four columns of \cref{tab:main_personalization}. The clearest separation is on the {PrefEval generation task, where a score of} $56.8$ is $13.2$ points above HumanLM, the strongest task-specific baseline in this split. In PersonaMem-v1 and PersonaMem-v2, memory augmentation through Mem0 improves the base model but remains $2.1$ and $2.2$ points below \ourmodel{}, respectively. {PrefEval classification scores are closer.} \ourmodel{} reaches $78.9$, compared to $78.5$ for HumanLM, so the $0.4$-point margin does not support a strong separation claim. \ourmodel{} exceeds GPT-4o-mini in three of the four evaluations, but remains below GPT-5-mini in all four. These results show that \ourcorpus{} improves a fixed backbone beyond {these baselines.}

\providecommand{\nd}{\textcolor{black!30}{--}}

\begin{table}[t]
\centering
\normalsize
\setlength{\tabcolsep}{4pt}
\renewcommand{\arraystretch}{1.06}
\caption{{\textbf{Personalization benchmark results.} \ourmodel{} leads the non-proprietary comparisons on all four metrics. All non-proprietary methods use Qwen2.5-7B-Instruct; entries are accuracy (\%).} \textbf{Bold} marks the best non-proprietary result in each column. {Dashes denote unreported results.}}
\label{tab:main_personalization}
\begin{tabular*}{\linewidth}{@{\extracolsep{\fill}}lrrrr@{}}
\toprule
\textbf{Model / Method}
 & \makecell{\textbf{{PersonaMem-v1}}\\[-1pt]\small MCQ}
 & \makecell{\textbf{{PersonaMem-v2}}\\[-1pt]\small MCQ}
 & \makecell{\textbf{PrefEval}\\[-1pt]\small {Generation}}
 & \makecell{\textbf{PrefEval}\\[-1pt]\small {Classification}} \\
\midrule
GPT-4o-mini & 48.6 & 37.3 & 21.1 & 84.6 \\
GPT-5-mini & 61.5 & 49.2 & 90.8 & 99.3 \\
\midrule
Qwen2.5-7B-Instruct & 49.9 & 32.0 & 23.4 & 62.0 \\
PersonaVLM & 33.9 & 24.3 & 29.8 & 54.7 \\
HumanLM & 38.1 & 27.3 & 43.6 & 78.5 \\
LLMoPt & 35.3 & 22.2 & 41.5 & 75.1 \\
Mem0 & 54.3 & 39.8 & \nd & \nd \\
\midrule
\textbf{\ourmodel{}} & \textbf{56.4} & \textbf{42.0} & \textbf{56.8} & \textbf{78.9} \\
\bottomrule
\end{tabular*}
\end{table}

\begin{table}[!b]
\centering
\small
\setlength{\tabcolsep}{4pt}
\renewcommand{\arraystretch}{1.06}
\caption{{\textbf{Belief and action reasoning results.} \ourmodel{} leads the Qwen2.5-7B-Instruct comparisons on BigToM, while AutoToM leads on ToMi. {Scores are percentages; parentheses give gains over the Qwen baseline in percentage points. BigToM requires correct answers to both true-belief and false-belief variants of each scenario.}} \textbf{Bold} marks the best non-proprietary result in each column.}
\label{tab:main_tom}
\begin{tabular*}{\linewidth}{@{\extracolsep{\fill}}lr@{\,}lr@{\,}lr@{\,}l@{}}
\toprule
\textbf{Model / Method}
 & \multicolumn{2}{c}{\textbf{ToMi}}
 & \multicolumn{2}{c}{\makecell{\textbf{{BigToM}}\\\textbf{{Forward Belief}}}}
 & \multicolumn{2}{c}{\makecell{\textbf{{BigToM}}\\\textbf{{Forward Action}}}} \\
\midrule
GPT-4o-mini & 77.8 & & 52.0 & & 78.0 & \\
GPT-5-mini & 89.5 & & 93.5 & & 88.0 & \\
\midrule
Qwen2.5-7B-Instruct & 80.5 & & 31.0 & & 23.5 & \\
\quad + AutoToM & \textbf{85.5} & {$(+5.0)$} & 40.3 & {$(+9.3)$} & 29.3 & {$(+5.8)$} \\
\quad + {ThoughtTracing} & 84.9 & {$(+4.4)$} & 35.0 & {$(+4.0)$} & 26.5 & {$(+3.0)$} \\
\midrule
\textbf{\ourmodel{}} & 82.3 & {$(+1.8)$} & \textbf{44.0} & {$(+13.0)$} & \textbf{31.0} & {$(+7.5)$} \\
\bottomrule
\end{tabular*}
\end{table}

\Needspace{6\baselineskip}
\noindent\textbf{{Improvements cover both response generation and answer selection.}}
{{\Cref{tab:response_format_scaling} compares generation with classification on Qwen's PrefEval tasks and with multiple-choice answering on Llama's PersonaMem-v2 tasks.}
On Qwen, PrefEval classification increases from $62.0$ to $78.9$, alongside the gain in preference-following generation.
On Llama, PersonaMem-v2 generation increases from $33.9$ to $45.6$, while multiple-choice accuracy rises from $23.9$ to $37.5$.
{The gains therefore extend to both answer selection and response generation based on user preferences.}}

{{Generation and answer selection follow different scaling trajectories.}
For Qwen, increasing the mixture from one quarter to one half raises PrefEval classification by $8.5$ points but generation by only $0.7$ points.
The final increase to the full mixture then raises generation by $18.9$ points and classification by $3.1$ points.
Llama's PersonaMem-v2 generation follows a different trajectory, reaching $44.0$ with one eighth of the data and $45.6$ with the full mixture.
Endpoint gains alone would conceal these differences in how the models benefit from additional supervision.}

\label{sec:tom_transfer}
\noindent\textbf{{\ourcorpus{} improves belief and action prediction on Qwen and Llama.}} In Qwen2.5-7B-Instruct, \ourmodel{} improves Forward Belief from $31.0$ to $44.0$ and Forward Action from $23.5$ to $31.0$. {Both gains exceed those of AutoToM and {ThoughtTracing}, while both baselines improve more on ToMi.} Transfer also scales with data on Llama-3.1-8B-Instruct (\cref{fig:scaling-zero-shot}). From Base to \textsc{All}, BigToM Forward Belief rises from $18.5$ to $43.3$, Forward Action from $39.5$ to $57.3$, and {ToMi accuracy increases} from $63.6$ to $70.6$.

\Needspace{5\baselineskip}
\noindent\textbf{{Llama improves across both domains with one quarter of the data.}}
{At the $\nicefrac{1}{4}$ scale (\cref{tab:sft_sample_scaling_unified}), Llama reaches $56.0$ on PrefEval generation, $36.4$ on BigToM Forward Belief, and $51.0$ on BigToM Forward Action.
These are gains of $31.5$, $17.9$, and $11.5$ percentage points over the base model.
At the same scale, ToMi improves from $63.6$ to $69.3$, and both PersonaMem-v2 response formats improve (\cref{tab:response_format_scaling}).
Increasing the mixture to full scale yields further gains on all these tasks, including another $9.4$ points on PrefEval generation and $6.9$ points on Forward Belief.
{Improvement across both domains is present at an intermediate data scale and continues as supervision increases.}}

\begin{table}[!t]
\centering
\normalsize
\setlength{\tabcolsep}{4pt}
\renewcommand{\arraystretch}{1.08}
\caption{{\textbf{Training gains across model families.} Personalization improves on every shared metric, while ToM gains depend on the backbone. Entries are full-mixture gains over the corresponding Instruct model in percentage points. PM denotes PersonaMem. Complete scores and training scales appear in \cref{tab:sft_sample_scaling_unified}.}}
\label{tab:cross_backbone_gains}
\begin{tabular*}{\linewidth}{@{\extracolsep{\fill}}lrrrrrr@{}}
\toprule
 & \multicolumn{3}{c}{\textbf{{Personalization}}}
 & \multicolumn{3}{c}{\textbf{{Theory of Mind}}} \\
\cmidrule(lr){2-4}\cmidrule(l){5-7}
\textbf{Backbone}
 & \makecell{\textbf{PrefEval}\\\small {Generation}}
 & \makecell{\textbf{PM-v1}\\\small MCQ}
 & \makecell{\textbf{PM-v2}\\\small MCQ}
 & \textbf{ToMi}
 & \makecell{\textbf{BigToM}\\\small {Forward Belief}}
 & \makecell{\textbf{BigToM}\\\small {Forward Action}} \\
\midrule
Qwen2.5-7B & $+33.4$ & $+6.5$ & $+10.0$ & $+1.8$ & $+13.0$ & $+7.5$ \\
Llama-3.1-8B & $+40.9$ & $+6.3$ & $+13.6$ & $+7.0$ & $+24.8$ & $+17.8$ \\
OLMo-3-7B & $+26.6$ & $+5.6$ & $+10.2$ & $+2.2$ & $-8.2$ & $-7.0$ \\
\bottomrule
\end{tabular*}
\end{table}

\begin{figure}[t]
  \centering
  \includegraphics[width=\linewidth]{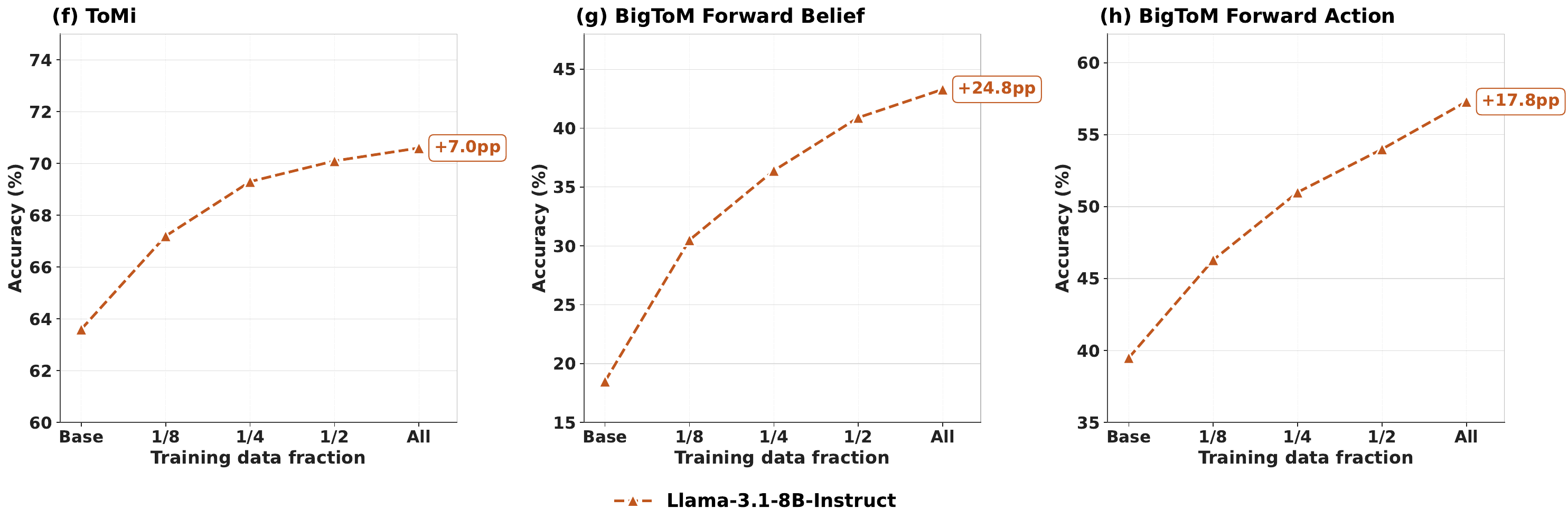}
  \caption{{\textbf{Belief and action prediction across training scales.} Llama-3.1-8B-Instruct improves on all three ToM tasks after training on \ourcorpus{}. Panels \textbf{(f)}, \textbf{(g)}, and \textbf{(h)} report accuracy on ToMi, BigToM Forward Belief, and BigToM Forward Action. {BigToM requires correct answers to both true- and false-belief variants.} Base is the unmodified backbone; fractions and All denote the amount of training data. Endpoint labels give gains in percentage points. \Cref{tab:sft_sample_scaling_unified} gives scores for all backbones.}}
  \label{fig:scaling-zero-shot}
\end{figure}

\begin{table}[!tb]
\centering
\normalsize
\setlength{\tabcolsep}{3.5pt}
\renewcommand{\arraystretch}{1.08}
\caption{{\textbf{Response generation and answer selection.} Both formats improve along different scaling trajectories. Entries report accuracy (\%) at each \ourcorpus{} scale. {\emph{Base} denotes the unmodified Instruct model; fractions specify shares of the full training mixture.} \Cref{tab:sft_sample_scaling_unified} gives full results.}}
\label{tab:response_format_scaling}
\begin{tabular*}{\linewidth}{@{\extracolsep{\fill}}lllrrrrr@{}}
\toprule
\textbf{{Backbone}} & \textbf{{Benchmark}} & \textbf{{Task}} & \textbf{Base}
 & $\nicefrac{1}{8}$ & $\nicefrac{1}{4}$ & $\nicefrac{1}{2}$ & \textbf{All} \\
\midrule
\multirow{2}{*}{{Qwen2.5-7B}}
 & \multirow{2}{*}{{PrefEval}} & {Generation} & 23.4 & 36.5 & 37.2 & 37.9 & 56.8 \\
 & & {Classification} & 62.0 & 63.5 & 67.3 & 75.8 & 78.9 \\
\addlinespace[2pt]
\multirow{2}{*}{{Llama-3.1-8B}}
 & \multirow{2}{*}{{PersonaMem-v2}} & {Generation} & 33.9 & 44.0 & 45.3 & 45.3 & 45.6 \\
 & & {MCQ} & 23.9 & 34.2 & 36.4 & 37.4 & 37.5 \\
\bottomrule
\end{tabular*}
\vspace{-10pt}
\end{table}

\Needspace{5\baselineskip}
\noindent\textbf{{Dialogue and QA supervision contribute complementary capabilities.}}
{The two corpus views teach models to respond within an interaction and to answer questions about the user.
To assess their contributions, we fine-tune Llama-3.1-8B-Instruct using dialogue examples alone, QA examples alone, or their full mixture (\cref{tab:supervision_view_ablation}).}

{Dialogue-only training improves the ToM macro average from $40.5$ to $46.3$ without MCQ supervision and achieves the highest scores on both generation tasks.
QA-only training achieves the highest PersonaMem-v2 MCQ score, but its generation scores fall below the unmodified model.
The full mixture reaches the highest ToM average ($57.1$) and retains generation performance close to dialogue-only training.
These results support combining response demonstrations with question-answer supervision to develop both assistance and mental-state reasoning.}

\Needspace{5\baselineskip}
\noindent\textbf{{Complementary evaluation distinguishes broad gains from task-specific effects.}}
{The backbone comparison reveals why personalization alone would give an incomplete account of human-aware learning (\cref{tab:cross_backbone_gains}).
OLMo gains $26.6$ points on preference-following generation and improves on both PersonaMem benchmarks and ToMi, yet loses $8.2$ and $7.0$ points on BigToM Forward Belief and Forward Action.
Qwen and Llama show gains in both domains, with Llama recording the largest improvements in belief and action prediction among the three backbones.

\Needspace{4\baselineskip}
{Evaluating both domains establishes benefits beyond personalized responses for two model families and identifies a concrete limit to that generality in the third.}
These findings make reasoning about people an empirical criterion for assistant training alongside the quality of the assistance itself.}

\begin{table}[!t]
\centering
\small
\color{black}
\setlength{\tabcolsep}{4pt}
\renewcommand{\arraystretch}{1.10}
\caption{{\textbf{Contributions of dialogue and QA supervision.} The full mixture gives Llama-3.1-8B-Instruct the highest ToM average; {dialogue-only training leads on generation, and QA-only training leads on answer selection.} Base is unmodified; ablations remove one view and change data volume. Scores are percentages, with column maxima in bold. \emph{ToM} averages ToMi, BigToM Forward Belief, and Forward Action equally.}}
\label{tab:supervision_view_ablation}
\begin{tabular*}{\linewidth}{@{\extracolsep{\fill}}lrrrr@{}}
\toprule
\textbf{Training condition}
 & \makecell{\textbf{PersonaMem-v2}\\\textbf{MCQ}}
 & \makecell{\textbf{ToM}\\\textbf{Macro average}}
 & \makecell{\textbf{PrefEval}\\\textbf{Generation}}
 & \makecell{\textbf{PersonaMem-v2}\\\textbf{Generation}} \\
\midrule
Base & 23.9 & 40.5 & 24.5 & 33.9 \\
Dialogue only & 24.8 & 46.3 & \textbf{66.0} & \textbf{45.8} \\
QA only & \textbf{38.6} & 52.3 & 22.7 & 31.7 \\
Full mixture & 37.5 & \textbf{57.1} & 65.4 & 45.6 \\
\bottomrule
\end{tabular*}
\vspace{-10pt}
\end{table}

\Needspace{6\baselineskip}
\vspace{-4pt}
\section{Conclusion}
\label{sec:conclusion}
\vspace{-4pt}

\begingroup\par\begingroup
\setlength{\parskip}{2pt}
{\oursys{} addresses the supervision gap in human-aware training by constructing interactions in which an assistant can directly observe the state of the user it serves.}
Shared-state user simulation gives the {\oracle{}} information about why a user acts, allowing it to demonstrate assistance informed by beliefs, goals, and circumstances that remain partly implicit in conversation.
\oursim{}, \ourcorpus{}, and \ourmodel{} implement this principle through controlled simulation, corpus construction, and {privileged} distillation with the state withheld from the student.

{Training on \ourcorpus{}} improves every reported personalization metric across three model families.
{The same supervision improves belief and action prediction on Qwen and Llama; OLMo's mixed results show that gains in personalized assistance can coexist with losses in mental-state reasoning.}
{User simulation therefore offers a way to design the experiences through which language models learn to understand the people they serve.}
{Future work should develop methods that actively elicit real users' intentions and identify useful supervision in everyday conversations, while protecting privacy and preserving users' control over training-data use.
Such evidence could guide more faithful simulators of how beliefs, goals, and circumstances evolve.
Combining real and simulated experience could advance AI collaborators that understand people's intentions and support their goals over time.}
\par\endgroup\endgroup

\par\endgroup
\bibliographystyle{mind2dialogue}
\begingroup
\renewcommand{\bibfont}{\small}
\setlength{\bibsep}{1.5pt plus 0.5pt minus 0.5pt}
\setlength{\parskip}{0pt}
\renewcommand{\bibsection}{%
  \vspace{-4pt}%
  \Needspace{60pt}%
  \section*{\refname}%
  \nobreak\vspace{-2pt}%
}
\bibliography{ref}
\endgroup

\newpage
\appendix

\section{Limitations}
\label{app:limitations}

\oursys{} studies synthetic supervision for a setting in which real long-horizon dialogue is difficult to collect. {The following limitations define the scope of the present evidence.}

\noindent \textbf{{Dependence on {\oracle{}} quality.}} All data-generation experiments use GPT-4o-mini as both the user simulator and the {\oracle{}} assistant. The corpus can therefore inherit this model's errors, stylistic biases, and limitations in representing user states. We do not test whether a stronger teacher, an ensemble, or a different simulator family would improve the student.

\noindent \textbf{Backbone-dependent Theory-of-Mind transfer.} {Qwen2.5-7B and Llama-3.1-8B improve on all three measured ToM tasks. OLMo-3-7B improves on ToMi but declines on both BigToM tasks.} For OLMo, the personalization gains persist (PrefEval-Gen $+26.6$, PersonaMem-v2 $+10.2$), while BigToM regresses (Forward Belief $-8.2$, Forward Action $-7.0$). Differences in pretraining, instruction tuning, optimization, or evaluation sensitivity could contribute to this result; the present experiments do not distinguish among them. {ToM transfer remains specific to the tested models and settings.}

\noindent\textbf{{Scope of capability evidence.}} {Our evaluation measures assistance and mental-state reasoning on separate benchmark suites, leaving their coordination within one interaction untested. The BigToM results cover Forward Belief and Forward Action. The supervision-view ablation varies both supervision content and data volume and does not isolate the student's reasoning process.}

\noindent \textbf{Latent-state schema.} The state document maintained by \oursim{} uses a hand-designed schema containing conversational context, stable attributes, transient attributes, and behavioral mode. This representation may omit relevant aspects of users or encode distinctions that do not transfer to real interactions. We do not compare alternative schemas or test sensitivity to individual state fields. Extending the schema may also change simulation dynamics and data quality, so compatibility with the remaining pipeline requires empirical validation.

\noindent \textbf{Validation against real long-horizon dialogue.} {Our evaluation combines public personalization and ToM benchmarks with LLM judgments and a human audit of synthetic conversations. These assessments do not establish how the trained students perform in sustained interaction with real users.} Longitudinal studies with informed consent and appropriate privacy protections are needed before drawing conclusions about real-user utility.

\section{Broader Impact}
\label{app:impact}

\noindent \textbf{Positive impact.} Personalization research often relies on private interaction logs or on synthetic conversations generated from static persona descriptions. \oursys{} provides an alternative source of training data in which both the simulated user and assistant are conditioned on a constructed state. The pipeline does not require collecting private {conversations between humans and assistants}, which reduces one source of privacy risk. Because the state and generation metadata are available to the data producer, the corpus can also support controlled analyses of persona, scenario, and behavioral coverage. A public release can facilitate replication and comparison, although synthetic generation does not by itself guarantee demographic balance, factual accuracy, or privacy safety.

\noindent \textbf{Potential negative impact and mitigation.} Systems designed to infer user goals, beliefs, or affect may also enable targeted persuasion, dependency, or unwanted profiling. Synthetic personas and state descriptions can encode stereotypes, and a model may express unwarranted confidence about mental states that are inherently uncertain. We do not evaluate manipulation, demographic bias, privacy leakage, or safety in mental-health-adjacent settings. Open release supports auditing but also broadens access to the capability. Any downstream deployment should therefore evaluate these risks directly, communicate uncertainty, provide user control over personalization and memory, and apply domain-appropriate safety and data governance measures. {\ourcorpus{} is released for research; production use requires further validation.}

\Needspace{12\baselineskip}
\section{Extended Related Work}
\label{app:related_work}
\label{app:extended-related-work}

{We organize the literature around four requirements for human-aware assistance, namely access to personal context, realistic interaction, reasoning about people, and supervision that remains useful when user states are unobserved.
The comparison below identifies what each research direction contributes and where shared-state supervision changes the learning problem.}

{%
\subsection{{Human-AI Collaboration and User Modeling}}}

{Sustained collaboration requires models to retain knowledge about people and use it as their intentions and circumstances change~\citep{collins2024people}.
Collaborative Gym provides shared task environments in which humans and agents communicate, coordinate actions, and retain control over their work~\citep{shao2026collaborative}.
LongMemEval tests information retention, reasoning across sessions, and knowledge updates, while HorizonBench evaluates preference tracking through simulated life events~\citep{wu2024longmemeval,li2026horizonbench}.
These studies motivate assistance that depends on an evolving understanding of the person.}

{Personalization studies how individual knowledge and preferences should affect model behavior within this broader collaboration problem.}
Recent benchmarks emphasize implicit preferences that are inferred from user behavior~\citep{zhao2025prefeval,jiang2025personamem,li2026horizonbench}.
{PersonaMem-v2 additionally studies how preference supervision can support reinforcement fine-tuning and agentic memory learning~\citep{jiang2025personamem}.}
{LaMP evaluates personalized classification and generation and studies retrieval from user profiles~\citep{salemi2024lamp}.
MemoryBank updates and retrieves memories from prior interactions, Mem0 extracts and consolidates salient conversational information, and long-term dialogue agents combine memory with user modeling~\citep{zhong2024memorybank,chhikara2025mem0,li2024ldagent}.
PLUM takes a parameter-based approach, augmenting previous conversations into QA examples for user-specific adapter training~\citep{magister2024plum}.
These approaches determine how personal information is retained and made available to the model.
\oursys{} constructs demonstrations of how personal context informs assistance.}
\looseness=-1

{Explicit user-state models extend personalization to reasoning about changes that are only partly observable in dialogue.
PUMA formulates interaction as decision-making under partial observability, maintains beliefs over user states, and selects actions using predicted state transitions~\citep{luo2026userstate}.
DreamCUB predicts future utterances and user beliefs in a dialogue world model and uses model-based reinforcement learning to improve the dialogue policy~\citep{zhao2025dream}.
Both methods make user-state reasoning part of assistant decision-making.
\oursys{} supplies a complementary source of supervision by constructing states during simulation and exposing them to the {\oracle{}} before generating training responses.
The student learns those responses with the evolving state withheld.}

{%
\subsection{Synthetic Dialogue and User Simulation}}

Persona-conditioned dialogue commonly starts from a fixed profile~\citep{zhang-etal-2018-personalizing}.
{Synthetic-Persona-Chat uses a generator and critics to expand persona-based conversations while checking their quality~\citep{jandaghi2024faithful}.
Persona Hub broadens the population available for data synthesis, while DeepPersona increases the depth and internal detail of synthetic profiles~\citep{ge2024scaling,wang2025deeppersona}.
Profile diversity and dialogue quality control are useful foundations, but a profile alone does not specify the sequence of mental states through which a user responds to an unfolding interaction.
\oursim{} adds an explicit state trajectory and uses each updated state to guide both user behavior and {\oracle{}} assistance.}

{Interactive simulation models how users pursue goals and respond to events over time.
Task-oriented user simulators support controlled dialogue evaluation~\citep{davidson2023user,sekulic2024reliable}, and UserLM trains the user role directly on human conversations conditioned on high-level intent~\citep{laban2025flipping}.
Generative Agents combine memory, reflection, and planning to produce behavior over time~\citep{park2023generative}.
LifeSim models long-horizon user lives and evolving intentions for personalized assistant evaluation~\citep{duan2026lifesim}.
HumanLM aligns latent states with real user responses, and OdysSim develops models trained across a broad collection of human behavior tasks~\citep{wu2026humanlm,zhou2026odyssim}.
These systems demonstrate that dynamic state and realistic behavior are established ingredients of simulation.}
We focus on \emph{who observes the simulator-defined user state when assistant responses are generated}.
In \oursim{}, the user simulator updates a structured state, and an {\oracle{}} assistant directly observes that same state when producing responses.

{Training through simulated interaction also requires deciding which aspect of the assistant's behavior receives supervision.
PersonaGym generates dynamic preference interactions, while its associated PPOpt method learns to rewrite prompts from inferred user profiles~\citep{ma2026synthetic}.
ProPerSim adapts a proactive assistant from user-specific feedback within a simulation of daily activities~\citep{kim2026propersim}.
Such feedback can improve assistants while preserving the separation between the user's internal state and the assistant's current estimate of it.
Our {\oracle{}} receives the simulator-defined state directly when constructing response targets.
This choice provides an explicit source of information for the demonstration, addressing the gap between generating a realistic user and generating an assistant response informed by that user.}

{%
\subsection{Social Intelligence and Mental-State Reasoning}}

{Human social cognition provides a basis for treating people's beliefs and goals as part of the problem an assistant must solve~\citep{baker2017mentalizing,collins2024people}.
Computational work on social intelligence examines both successful interaction and the reasoning needed to interpret other agents.
SOTOPIA-$\pi$ develops social behavior through behavior cloning and self-reinforcement on evaluated interactions~\citep{wang2024sotopia}.
Its learning signal concerns the quality of the interaction; \oursys{} additionally specifies the user state available to the assistant that generates supervision.
This distinction concerns how demonstrations are constructed and can be combined with improvements in interaction-based learning.}

Theory-of-Mind (ToM) benchmarks test whether models track beliefs and other mental states in narratives and conversations~\citep{le2019revisiting,gandhi2023bigtom,kim-etal-2023-fantom,chen2024tombench,xu-etal-2024-opentom}.
{FANToM makes information asymmetry central to its conversational evaluation, while OpenToM and ToMBench broaden the range of characters, states, and questions.}
ToMATO couples characterized speakers with verbalized mental states~\citep{shinoda2025tomato}.
{Its agents' thoughts, goals, and personalities are hidden from their partners, and the thoughts supply mental-state QA labels.
ToMATO also studies fine-tuning on separately generated QA data.
\oursys{} exposes the user's evolving state to an {\oracle{}} that generates assistant responses, which then supervise a student without state access.}
\looseness=-1

{Joint evaluation on personalization and ToM tests two consequences of training models to attend to people.
Personalization measures whether the assistant uses information about the user when responding; ToM measures whether the same training benefits explicit reasoning about another agent's beliefs and actions.
The two suites provide separate behavioral observations, and sensitivity to perturbations remains relevant when interpreting ToM scores~\citep{ullman2023large,shapira2024clever}.
The mixed BigToM results across backbones further delimit the empirical connection reported in this paper.}

{%
\subsection{Learning with Privileged Information}}

{Learning with privileged information allows training to benefit from information unavailable at test time~\citep{vapnik2009new}.
Knowledge distillation transfers a teacher's predictions~\citep{hinton2015distilling,wang2023multitask}, and generalized distillation connects this principle to teachers and students receiving different representations of an example~\citep{lopez-paz2016unifying}.
This literature motivates our use of {an \oracle{} with access to the simulated user state} to supervise an assistant that cannot directly observe the evolving user state.}

{Recent work studies how to optimize this transfer for language models.
Privileged Information Distillation jointly trains conditioned teachers and students with shared parameters, and its on-policy variant uses a conditioned teacher to regularize student behavior~\citep{penaloza2026privileged}.
}
By contrast, our {\oracle{}} is frozen; the teacher and student use separate parameters; and transfer is supervised.
{Shared-state simulation supplies the additional information and links it to user behavior before distillation begins.
The resulting contribution is a construction of supervision for human-aware assistance that can operate with standard language model training.}

\newpage
\section{{Complete Evaluation Results and Simulation Analysis}}
\label{sec:full_experiments}

{We report the complete student scaling results underlying \cref{sec:main_results}, followed by the simulator validation and component ablation supporting \cref{sec:preliminary}.}

\subsection{{Complete Data Scaling Results}}
\label{app:full_scaling}

{\Cref{tab:sft_sample_scaling_unified} retains every reported score and training fraction. The main-text summary in \cref{tab:cross_backbone_gains} selects full-mixture gains on metrics available for all three backbones.}

\newcommand{\sftgain}[1]{{\scriptsize\,(+#1)}}
\newcommand{\sftloss}[1]{{\scriptsize\,(-#1)}}
\newcommand{\sftna}{\textcolor{black!45}{--}}

\begin{table}[H]
    \centering
 \caption{{\textbf{Complete results across training scales.} Full-mixture training improves every reported personalization metric; ToM outcomes vary across backbones. Panels \textbf{(a)} and \textbf{(b)} report personalization and mental-state reasoning, respectively.} All entries are accuracy (\%). {PM denotes PersonaMem; BigToM requires correct answers to both true- and false-belief variants.} \textbf{Bold} marks the best score within each backbone and metric; {parentheses give full-mixture changes from the base model in percentage points.} {Base denotes the unmodified model, fractions specify the training-mixture size, and dashes indicate unreported results.}}
    \label{tab:sft_sample_scaling_unified}

    \footnotesize
    \setlength{\tabcolsep}{3pt}
    \renewcommand{\arraystretch}{1.08}

    \textbf{(a) Personalization benchmarks}\par\smallskip
    \begin{tabularx}{\linewidth}{@{}>{\raggedright\arraybackslash}p{0.19\linewidth}l*{5}{>{\centering\arraybackslash}X}@{}}
        \toprule
        \textbf{Backbone}
        & \textbf{Scale}
        & \makecell[c]{\textbf{PrefEval}\\[-1pt]\scriptsize Gen.}
        & \makecell[c]{\textbf{PrefEval}\\[-1pt]\scriptsize Cls.}
        & \makecell[c]{\textbf{PM-v1}\\[-1pt]\scriptsize MCQ}
        & \makecell[c]{\textbf{PM-v2}\\[-1pt]\scriptsize MCQ}
        & \makecell[c]{\textbf{PM-v2}\\[-1pt]\scriptsize Gen.} \\
        \midrule
        \textbf{Qwen2.5-7B}
        & Base & 23.4 & 62.0 & 49.9 & 32.0 & \sftna \\
        & SFT-$\nicefrac{1}{8}$ & 36.5 & 63.5 & 52.8 & 34.5 & \sftna \\
        & SFT-$\nicefrac{1}{4}$ & 37.2 & 67.3 & 54.2 & 38.3 & \sftna \\
        & SFT-$\nicefrac{1}{2}$ & 37.9 & 75.8 & 55.3 & 38.5 & \sftna \\
        & SFT-All
        & \textbf{56.8}\sftgain{33.4}
        & \textbf{78.9}\sftgain{16.9}
        & \textbf{56.4}\sftgain{6.5}
        & \textbf{42.0}\sftgain{10.0}
        & \sftna \\
        \midrule
        \textbf{Llama-3.1-8B}
        & Base & 24.5 & \sftna & 42.1 & 23.9 & 33.9 \\
        & SFT-$\nicefrac{1}{8}$ & 38.3 & \sftna & 44.8 & 34.2 & 44.0 \\
        & SFT-$\nicefrac{1}{4}$ & 56.0 & \sftna & 45.7 & 36.4 & 45.3 \\
        & SFT-$\nicefrac{1}{2}$ & 61.6 & \sftna & 46.2 & 37.4 & 45.3 \\
        & SFT-All
        & \textbf{65.4}\sftgain{40.9}
        & \sftna
        & \textbf{48.4}\sftgain{6.3}
        & \textbf{37.5}\sftgain{13.6}
        & \textbf{45.6}\sftgain{11.7} \\
        \midrule
        \textbf{OLMo-3-7B}
        & Base & 52.5 & \sftna & 37.2 & 30.1 & \sftna \\
        & SFT-$\nicefrac{1}{8}$ & 61.8 & \sftna & 37.4 & 36.0 & \sftna \\
        & SFT-$\nicefrac{1}{4}$ & 68.7 & \sftna & 41.8 & 37.9 & \sftna \\
        & SFT-$\nicefrac{1}{2}$ & 75.3 & \sftna & 41.8 & 39.8 & \sftna \\
        & SFT-All
        & \textbf{79.1}\sftgain{26.6}
        & \sftna
        & \textbf{42.8}\sftgain{5.6}
        & \textbf{40.3}\sftgain{10.2}
        & \sftna \\
        \bottomrule
    \end{tabularx}

    \medskip
    \textbf{(b) Theory-of-Mind transfer benchmarks}\par\smallskip
    \begin{tabularx}{\linewidth}{@{}>{\raggedright\arraybackslash}p{0.24\linewidth}l*{3}{>{\centering\arraybackslash}X}@{}}
        \toprule
        \textbf{Backbone}
        & \textbf{Scale}
        & \textbf{ToMi}
        & \makecell[c]{\textbf{BigToM}\\[-1pt]\scriptsize Forward Belief}
        & \makecell[c]{\textbf{BigToM}\\[-1pt]\scriptsize Forward Action} \\
        \midrule
        \textbf{Qwen2.5-7B}
        & Base & 80.5 & 31.0 & 23.5 \\
        & SFT-$\nicefrac{1}{8}$ & \sftna & 36.5 & \sftna \\
        & SFT-$\nicefrac{1}{4}$ & \sftna & 39.5 & \sftna \\
        & SFT-$\nicefrac{1}{2}$ & \sftna & 42.0 & \sftna \\
        & SFT-All & \textbf{82.3}\sftgain{1.8}
        & \textbf{44.0}\sftgain{13.0}
        & \textbf{31.0}\sftgain{7.5} \\
        \midrule
        \textbf{Llama-3.1-8B}
        & Base & 63.6 & 18.5 & 39.5 \\
        & SFT-$\nicefrac{1}{8}$ & 67.2 & 30.5 & 46.3 \\
        & SFT-$\nicefrac{1}{4}$ & 69.3 & 36.4 & 51.0 \\
        & SFT-$\nicefrac{1}{2}$ & 70.1 & 40.9 & 54.0 \\
        & SFT-All & \textbf{70.6}\sftgain{7.0}
        & \textbf{43.3}\sftgain{24.8}
        & \textbf{57.3}\sftgain{17.8} \\
        \midrule
        \textbf{OLMo-3-7B}
        & Base & 78.1 & \textbf{30.2} & \textbf{59.2} \\
        & SFT-$\nicefrac{1}{8}$ & \sftna & \sftna & \sftna \\
        & SFT-$\nicefrac{1}{4}$ & \sftna & \sftna & \sftna \\
        & SFT-$\nicefrac{1}{2}$ & \sftna & \sftna & \sftna \\
        & SFT-All & \textbf{80.3}\sftgain{2.2}
        & 22.0\sftloss{8.2}
        & 52.2\sftloss{7.0} \\
        \bottomrule
    \end{tabularx}
\end{table}

\subsection{Simulator Validation and Component Ablations}
\label{app:pilot_tables}

{\oursim{} receives higher trajectory-level judge scores than Vanilla across all five measured dimensions. We evaluate $20$ held-out personas using the LLM-as-judge protocol of \citet{luo2026spasmstablepersonadrivenagent}. The composite $Z$-score gap is $1.15$ ($d{=}2.12$), and \oursim{} receives the higher aggregate score for all $20$ paired personas (binomial $p{<}0.001$). \Cref{tab:sim_quality} reports all scores and effect sizes.}

\newcommand{\best}[1]{\textbf{#1}}
\newcommand{\posd}[1]{{#1}}

\begin{table}[H]
    \centering
    \small
    \caption{{\textbf{Simulator validation on held-out personas.} \oursim{} scores higher than Vanilla across all five judge dimensions on $20$ held-out personas. Vanilla omits state maintenance and behavior guidance. Compare methods within each row because score scales differ; the composite row reports standardized scores. Positive Cohen's $d$ favors \oursim{}. \textbf{Bold} marks the higher score.}}
    \label{tab:sim_quality}
    \setlength{\tabcolsep}{7pt}
    \renewcommand{\arraystretch}{1.08}
    \begin{tabularx}{\linewidth}{@{}Xrrr@{}}
        \toprule
        \textbf{Dimension}
        & \textbf{{\oursim{}}}
        & \textbf{Vanilla}
        & Cohen's $d$ \\
        \midrule
        {\oracle{}} Personalization      & \best{3.28} & 2.97 & \posd{0.87} \\
        Information Efficiency      & \best{0.32} & 0.24 & \posd{2.05} \\
        Cross-Persona Distinctness  & \best{0.87} & 0.83 & \posd{1.08} \\
        Turn Novelty                & \best{0.78} & 0.69 & \posd{1.19} \\
        Topic Depth                 & \best{3.72} & 2.72 & \posd{1.00} \\
        \midrule
        \textbf{Composite Z-score}
        & {$\mathbf{+0.57}$}
        & $-$0.58
        & \posd{\textbf{2.12}} \\
        \bottomrule
    \end{tabularx}
\end{table}

\noindent\textbf{State maintenance supports persona specificity.} A component ablation varies state maintenance and behavior prompting independently (\cref{tab:m2d_component_ablation}). Compared to the full simulator, conditions without the state document have lower persona-specificity scores at turn~20 (\emph{Vanilla}: $d{=}{-}0.66$, $p{<}.01$; \emph{Profile-only Oracle}: $d{=}{-}0.49$, $p{<}.05$). The two conditions decline over the evaluated horizon, whereas the state-maintaining conditions increase. Vanilla user messages are $2.4\times$ longer on average, suggesting that verbosity alone does not explain the observed gap.

\begin{table}[H]
  \centering
  \small
  \setlength{\tabcolsep}{6pt}
  \renewcommand{\arraystretch}{1.2}
  \newcommand{\reddown}{\,$\downarrow$}

 \caption{{\textbf{Ablation of state and behavior controls.} The full simulator has the highest persona specificity at turn $20$. {\emph{State}} and {\emph{Behavior}} indicate state maintenance and behavior guidance. Scores track user persona specificity at initialization and turn $20$; negative $d$ denotes lower final specificity than \oursim{}.} Persona specificity evaluates the simulated user trajectory and therefore does not directly measure the {\oracle{}}'s access to user state. {Downward arrows mark declines since initialization.}}
  \label{tab:m2d_component_ablation}

  \begin{tabular*}{\linewidth}{@{\extracolsep{\fill}}lccccc@{}}
  \toprule
    & \multicolumn{2}{c}{\textbf{Components}}
    & \multicolumn{2}{c}{\textbf{Persona specificity (1--5)}}
    & \\
  \cmidrule(lr){2-3} \cmidrule(lr){4-5}
  \multirow{-2}{*}{\textbf{Condition}}
    & \textbf{State} & \textbf{Behavior}
    & \textbf{$t{=}0$} & \textbf{$t{=}20$}
    & \multirow{-2}{*}{\textbf{\makecell{$d$ vs.\\[-2pt] \oursim{}}}} \\
  \midrule
  Vanilla
    & \xmark & \xmark & 3.31 & 3.20\reddown & $-0.66^{**}$ \\
  Profile-only {\oracle{}}
    & \xmark & \cmark & 3.35 & 3.28\reddown & $-0.49^{*\phantom{*}}$ \\
  \addlinespace[4pt]
  Stateful (no behavior controller)
    & \cmark & \xmark & 3.26 & 3.35 & $-0.15^{\phantom{**}}$ \\
  \addlinespace[4pt]
  \textbf{\oursim{}}
    & \cmark & \cmark & 3.32 & \textbf{3.40} & ref. \\
  \bottomrule
  \end{tabular*}

  \smallskip
  \begin{minipage}{0.92\linewidth}
  \footnotesize
  \textit{Note.} Vanilla {user messages} are $2.4\times$ longer than \oursim{} {user messages}
  (97.6 vs.\ 40.6 words per message), {suggesting that verbosity alone does not explain the gap.}
  {The effect is largest for} \textit{goal} ($d{=}1.03$){; effects on}
  \textit{identity} ($d{=}{-}0.05$) {and} \textit{communication}
  ($d{=}{-}0.08$) {are near zero.} $^{*}p{<}.05$; $^{**}p{<}.01$.
  \end{minipage}
\end{table}

\newpage
\section{Simulation Algorithm}

{\Cref{alg:m2dsim} details \oursim{} and corpus construction (\cref{sec:ours,sec:corpus_construction}).}

\begin{algorithm}[H]
  \caption{{\textbf{Shared-state corpus generation.}}}
  \label{alg:m2dsim}
  \small
  \begin{algorithmic}[1]
    \Require Persona $p$; Scenario $\mathcal{S}_p$; Behavior Mode Families $\mathcal{F}{=}\{F_1,\dots,F_6\}$; Horizon $T$; Threshold $\theta_{\mathrm{sim}}$
    \Ensure Corpus $\mathcal{D}$, State Trajectory $s_{1:T}$
    \Statex \hspace{-1.2em}\emph{Scenario sampling}
    \Repeat
    \State $\sigma \sim \mathrm{Unif}\{\textsc{Lifelong}, \textsc{HighFreq}, \textsc{Affective}\}$
    \State $x \gets \textsc{ProposeScenario}(p, \sigma)$
    \Until{{$x$ passes the persona-consistency and abstraction checks and the distinctness check at threshold $\theta_{\mathrm{sim}}$ (\cref{sec:ours})}}
    \State $\mathcal{S}_p \gets \mathcal{S}_p \cup \{x\}$;\quad $s_0 \gets \textsc{InitState}(p, x)$;\quad $H_{<1} \gets \varnothing$;\quad $\mathcal{D} \gets \varnothing$
    \Statex \hspace{-1.2em}\emph{Rollout}
    \For{$t = 1$ \textbf{to} $T$}
    \State $s_t \gets f_{\mathrm{state}}(p, s_{t-1}, H_{<t})$ \Comment{unresolved goals, affect, trust}
    \State $F \gets \textsc{SelectFamily}(\mathcal{F}, \text{coverage}_{<t})$
    \State $b_t \sim \mathrm{Unif}(F)$;\quad $g_t \gets \textsc{Guidance}(b_t, t)$
    \State $m_t \gets f_{\mathrm{user}}(p, s_t, H_{<t}, g_t)$ \label{line:user}
    \State $a_t \gets f_{\mathrm{oracle}}(p, s_t, H_{<t}, m_t)$ \label{line:oracle}
    \State $H_{<t+1} \gets H_{<t} \,\Vert\, (m_t, a_t)$
    \State $\mathcal{D} \gets \mathcal{D} \cup \{(p, (H_{<t}, m_t), a_t)\}$
    \EndFor
    \Statex \hspace{-1.2em}\emph{Retention}
    \If{$H_{<T+1}$ passes the four programmatic and two judge checks}
    \State \Return $\mathcal{D} \cup \textsc{MakeQA}(p, s_{1:T}, H_{<T+1})$
    \Else
    \State \textbf{discard} the trajectory
    \EndIf
  \end{algorithmic}
\end{algorithm}

\section{Dataset Statistics}
\label{sec:dataset_statistics}

The public release of \ourcorpus{} comprises samples drawn from $289$ personas: deep-scenario multi-turn conversations, a broader rollout corpus, and QA-format training items (\cref{app:stats_mixture} gives the composition of the subset used for training). This section characterizes the $6{,}330$-conversation deep-scenario subset, whose conversations are paired with full latent-state trajectories. Its personas come from five focal countries (U.S., China, Japan, Germany, and India), whereas the broader rollout corpus additionally uses a global persona pool.

\subsection{Scenario coverage}
\label{app:stats_scenarios}

All category statistics below are read directly from the \texttt{scenario\_category} field assigned at generation time. \Cref{sec:corpus_coverage} gives family-level counts; this section reports them by category.

\begin{figure}[!t]
  \centering
  \includegraphics[width=\linewidth]{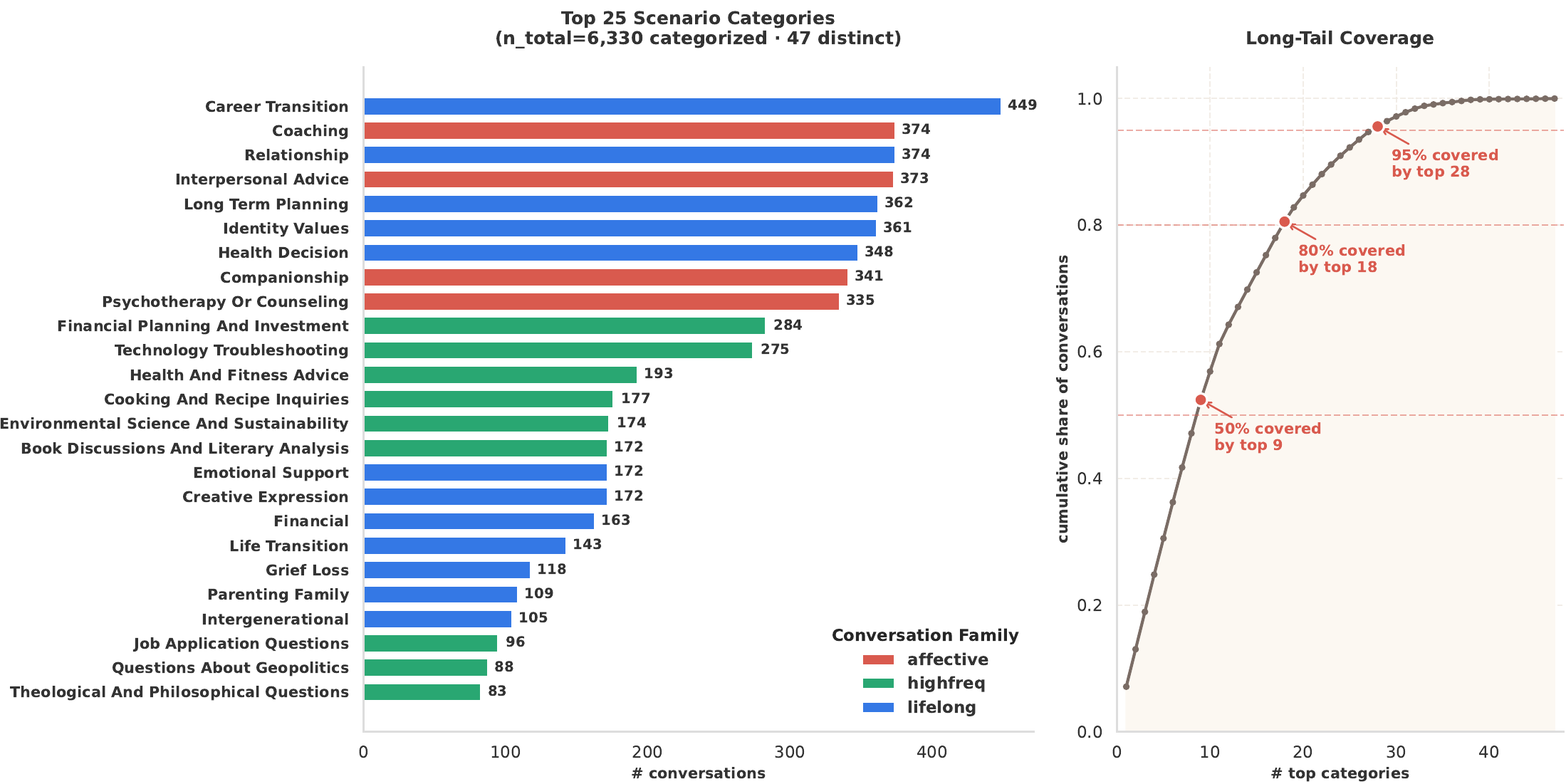}
  \caption{{\textbf{Scenario-category distribution.} Scenario frequencies are uneven across the $47$ categories in the $6{,}330$-conversation subset. The left panel shows counts for the $25$ largest categories, colored by scenario family. The right panel shows cumulative conversation coverage as categories are added in descending frequency.}}
  \label{fig:scenario_coverage}
\end{figure}

{\Cref{fig:scenario_coverage} shows the $25$ largest categories (left) and cumulative coverage of all $47$ categories (right). {The left panel in \cref{fig:dataset_composition}} displays category shares and labels those of at least $2.7\%$.}

\subsection{Persona specializations}
\label{app:stats_specializations}

{The persona pool contains $153$ distinct free-form \texttt{specialization} values. We embed these strings with all-MiniLM-L6-v2~\citep{reimers2019sentence} and apply $K$-means with $K=10$ and random seed $42$. Each cluster is labeled by the specialization associated with the most conversations. {The right panel in \cref{fig:dataset_composition}} counts one item per conversation and therefore shows the contribution of each specialization cluster {within the analyzed conversation subset.}}

\Needspace{7\baselineskip}
\subsection{Alignment to profile and scenario}
\label{app:stats_alignment}

\begin{figure}[H]
  \centering
  \includegraphics[width=\linewidth]{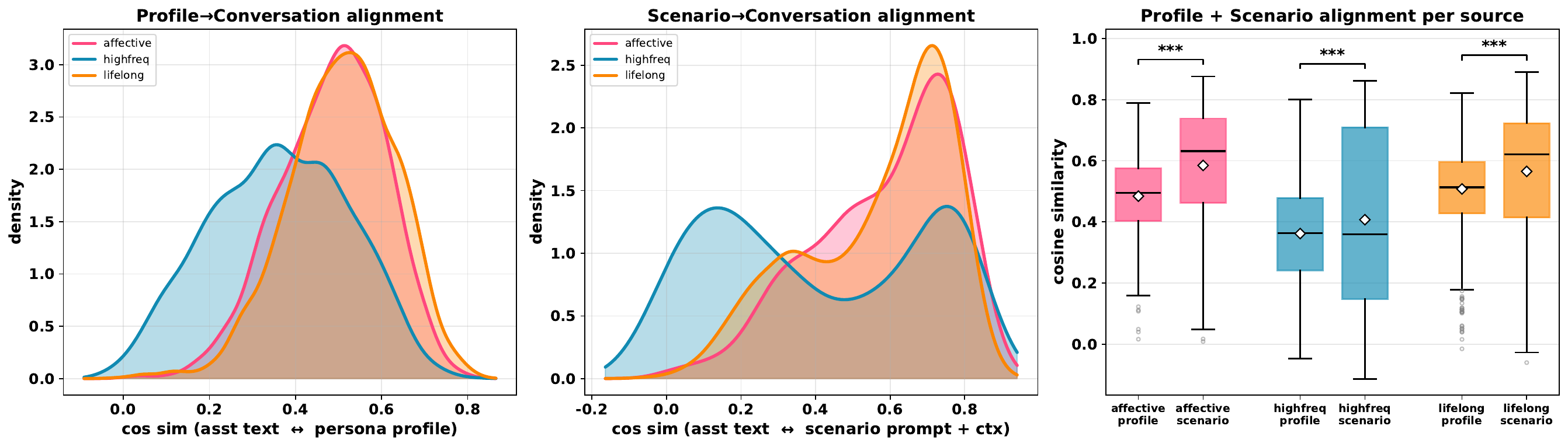}
  \caption{{\textbf{Alignment with personas and scenarios.} {Assistant text is more similar to scenarios than to personas in each source. \emph{Left and middle:} Cosine similarity distributions for persona and scenario references.} {\emph{Right:}} Paired comparisons, with black bars for medians and white diamonds for means. The subsets contain $471$ affective, $897$ high-frequency, and $2{,}187$ lifelong conversations; $^{***}p<0.001$ in Wilcoxon signed-rank tests.}}
  \label{fig:conv-alignment}
\end{figure}

For each conversation included in the alignment analysis, we compute the cosine similarity between the assistant text and two references: the static persona and the scenario prompt together with its context. {The left and middle panels in \cref{fig:conv-alignment}} show kernel density estimates of the two distributions for each data source. {The right panel} shows paired values within each source, with the median drawn as a black bar and the mean as a white diamond.

The two references are not interchangeable. Within each analyzed source subset ($471$ affective, $897$ highfreq, and $2{,}187$ lifelong), a paired Wilcoxon signed-rank test finds greater alignment with the scenario than with the static persona ($p<0.001$ in all three cases).

\Needspace{8\baselineskip}
\subsection{Behavioral modes}
\label{app:stats_modes}

\begin{figure}[H]
  \centering
  \includegraphics[width=\linewidth]{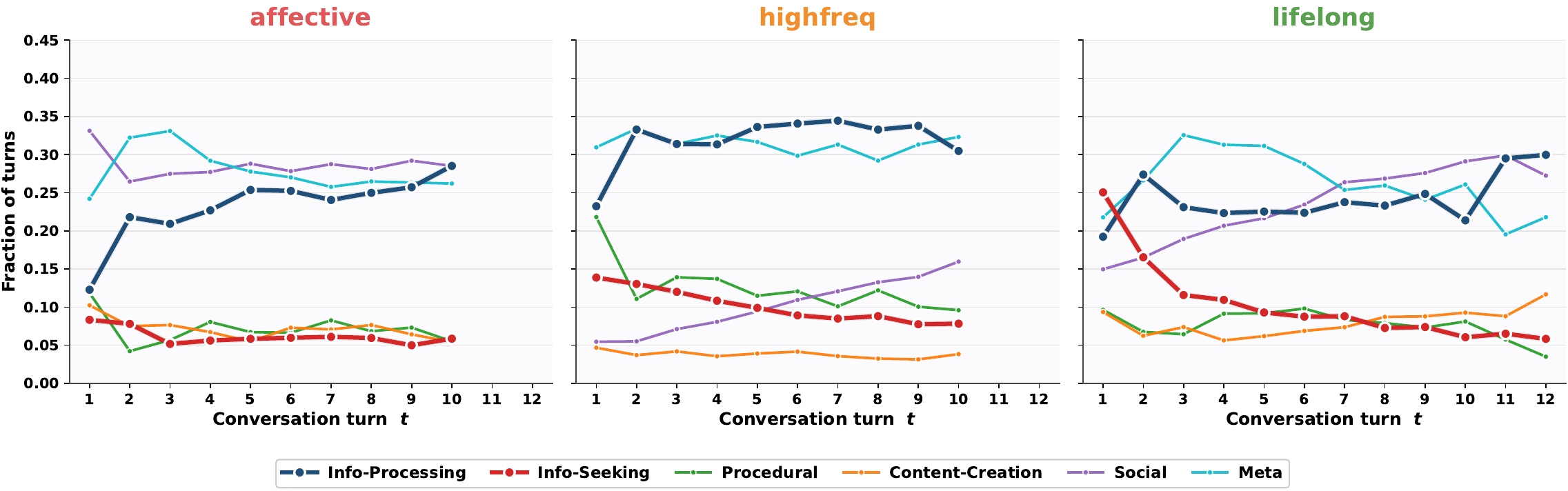}
  \caption{{\textbf{Behavioral composition across dialogue turns.} Behavior-family shares vary with turn position and conversation source. Each panel reports family proportions within a source, normalized after excluding fallback modes. Colors identify the six families. Points are shown through turn $12$ only when at least five retained turns are available for that source and position.}}
  \label{fig:mode-family-shift}
\end{figure}

\begingroup\par\begingroup
The behavioral analysis classifies the user turns produced by the simulator, because controller selections were not retained as labels in the generated conversations.

\Needspace{8\baselineskip}
For each of the $16$ {controller} modes, we form a prototype $\pi_m$ by concatenating the mode description with up to five example user turns from the {mode catalog}.
Let $\phi(\cdot)\in\mathbb{S}^{d-1}$ be the $\ell_2$-normalized all-MiniLM-L6-v2 embedding, with $d=384$.
Each turn $u$ receives the label
\begin{equation}
  \hat{m}(u)=\operatorname*{arg\,max}_{m\in\mathcal{M}}
  \langle\phi(u),\phi(\pi_m)\rangle,
  \qquad |\mathcal{M}|=16.
\end{equation}
The inner product is cosine similarity, so each label identifies the nearest mode prototype.

\Cref{fig:mode-distribution} summarizes $61{,}176$ turns across the $14$ displayed modes.
The inner ring shows six mode families, the middle ring separates individual modes, and the outer ring shows the contribution of each conversation source within a mode.
\Cref{fig:mode-family-shift} uses turn positions to describe how family proportions change across the $65{,}107$ user turns in the $6{,}330$-conversation analysis corpus.
We normalize each source and turn position after excluding fallback modes.
\par\endgroup\endgroup Formally, a surjection $F:\mathcal{M}\to\mathcal{F}\cup\{\textsf{Other}\}$ maps the $16$ modes to the six functional families $\mathcal{F}$; the two fallback modes (\texttt{compound\_request} and \texttt{default\_behavior}) are mapped to \textsf{Other} and excluded before renormalization. Letting $U_{s,t}$ denote the set of user turns at position $t$ from source $s$, the plotted family proportion is
\begin{equation}
  p_{s,t}(f) \;=\;
  \frac{\bigl|\{\, u \in U_{s,t} \;:\; F(\hat{m}(u)) = f \,\}\bigr|}
       {\bigl|\{\, u \in U_{s,t} \;:\; F(\hat{m}(u)) \neq \textsf{Other} \,\}\bigr|},
  \qquad f \in \mathcal{F},
\end{equation}
such that $\sum_{f\in\mathcal{F}} p_{s,t}(f)=1$ for every {plotted} $(s,t)$. {We restrict the analysis to $t\leq12$ and plot a point only when at least five turns remain after excluding \textsf{Other} for the corresponding source and turn position.}

\subsection{SFT mixture composition}
\label{app:stats_mixture}

The models evaluated in this paper are trained on a subset of the corpus described above. We derive training examples from two views of the same latent-state trajectories. In the {dialogue} view, the student receives the persona and the observable {dialogue} history, with the state-aware {\oracle{}} response as the target. In the QA view, the student receives the observable history, a generated question, and answer options when required by the task format. Neither view exposes the latent-state document $s_t$ to the student. {Only the user simulator, {\oracle{}}, and QA generator can access this document.}

\begin{table}[htbp]
  \centering
  \small
  \caption{{\textbf{Composition of the training mixture.} The $8{,}244$ examples combine dialogue demonstrations with three QA formats.} All components are derived from \oursim{} trajectories that passed quality control. The mixture contains no instances, histories, questions, or answer options from the evaluation benchmarks.}
  \label{tab:sft_mixture}
  \begin{tabularx}{\linewidth}{@{}lXr@{}}
    \toprule
    \textbf{View}                & \textbf{Component}                  & \textbf{Examples} \\
    \midrule
    Dialogue                     & Multi-turn conversation supervision & $3{,}312$         \\
    QA                           & PersonaMem-format MCQ               & $2{,}052$         \\
    QA                           & Open-ended preference QA            & $1{,}442$         \\
    QA                           & Preference-classification QA        & $1{,}438$         \\
    \midrule
    \multicolumn{2}{@{}l}{Total} & $8{,}244$                                               \\
    \bottomrule
  \end{tabularx}
\end{table}

The QA components provide training examples in the multiple-choice and classification formats used by several evaluations. The {dialogue component provides} free-form targets generated by an {\oracle{}} that has access to the latent state. One QA component follows the four-option PersonaMem interface. We manually specify the generation procedure for this component, while \oursim{} generates each persona, scenario, trajectory, question, and answer option. Training and evaluation data are therefore matched in the interface but disjoint in content (\cref{sec:setup}). Performance under this matched interface provides evidence within that setting, but does not by itself establish transfer to formats not represented during training.

\Needspace{9\baselineskip}
\subsection{{Training and Evaluation Data Provenance}}
\label{app:data_provenance}

{The evaluation benchmarks are held out from corpus construction.
The simulator, {\oracle{}}, and QA constructor do not condition on benchmark examples, histories, questions, or answer options.
We compare the released \ourcorpus{} with PersonaMem-v1, PersonaMem-v2, PrefEval, ToMi, and BigToM using exact 13-gram matching (\cref{tab:benchmark_overlap}).
No matches are detected across the five suites.}

\begin{table}[htbp]
\centering
\small
\color{black}
\setlength{\tabcolsep}{5pt}
\renewcommand{\arraystretch}{1.12}
\caption{{\textbf{Exact overlap with evaluation benchmarks.} Comparing the released \ourcorpus{} with all five evaluation suites finds zero matching 13-grams. Each row identifies the benchmark, its evaluation target, and the number of exact matches detected.}}
\label{tab:benchmark_overlap}
\begin{tabularx}{\linewidth}{@{}lXr@{}}
\toprule
\textbf{Benchmark} & \textbf{Evaluation target} & \makecell[r]{\textbf{Exact 13-gram}\\\textbf{matches}} \\
\midrule
PersonaMem-v1 & Memory and evolving user profiles & 0 \\
PersonaMem-v2 & Implicit user preferences & 0 \\
PrefEval & Preference adherence & 0 \\
ToMi & First- and second-order beliefs & 0 \\
BigToM & Belief and action prediction & 0 \\
\bottomrule
\end{tabularx}
\end{table}

{The overlap audit concerns exact text reuse between the generated corpus and the evaluation benchmarks.
The check cannot exclude paraphrased overlap or benchmark exposure during teacher or student pretraining.
PersonaMem MCQ also shares a response interface with one training component, so its gains are interpreted under that matched format (\cref{app:stats_mixture}).}

\Needspace{8\baselineskip}
\subsection{Quality control}
\label{app:qc}

Before inclusion in \ourcorpus{}, each conversation generated by \oursim{} undergoes six quality checks. Four programmatic checks examine properties that can be determined from the conversation record. Two LLM-judge checks assess whether the simulated user remains consistent with the assigned persona. \Cref{app:qc_prompts} provides the complete rubrics for both LLM-judge checks.

{The programmatic checks verify schema validity; turn count, role alternation, and token bounds; state-trajectory completeness; and binding to the assigned profile.
A conversation enters the retained corpus only after passing all six checks.}

\noindent\textbf{LLM judge dimensions.}
\begin{itemize}[leftmargin=1.4em,itemsep=2pt,topsep=2pt]
  \item \textbf{Persona consistency}. The judge assigns a Likert score from $1$ to $5$ based on how consistently the simulated user's messages reflect the assigned profile and behavioral metadata. The judge also provides a written justification.

  \item {\textbf{Profile contradiction.} The judge labels each conversation \texttt{no\_contradiction}, \texttt{unclear}, or \texttt{contradicts}, according to whether a user turn conflicts with an immutable profile attribute. For a detected conflict, the judge also records the relevant turn index.}
\end{itemize}

\Needspace{10\baselineskip}
\noindent\textbf{Human audit.}
We conducted a human audit to estimate the quality of conversations retained after automatic filtering. We uniformly sampled $1{,}240$ conversations that had passed all six checks. Annotators evaluated each conversation using a binary rubric that covers persona consistency, trajectory coherence, state-response consistency, and response relevance. Of the {sampled} conversations, $1{,}216$ satisfied the rubric ($98.06\%$). Most of the remaining cases involved prompt seed capture, in which a generic scenario displaced the assigned persona.
{The $24$ conversations that failed the rubric represent a residual violation rate of $1.94\%$ in this post-filtering sample.
Human review measured the quality of retained supervision and did not serve as a corpus-wide manual filtering stage.}

\section{Prompts}
\label{sec:prompt_templates}

\subsection{Scenario Generation}
\label{app:scenario_prompts}

\subsubsection{Lifelong Scenario}
\label{app:prompt_lifelong}

\begin{promptbox}
  \begin{Verbatim}[fontsize=\scriptsize,breaklines=true,breakanywhere=true,breakautoindent=false,breakindent=0pt,breaksymbolleft={},breaksymbolsepleft=0pt]
You are creating deeply personal conversation-starting messages 
for a specific person in a conversation with an AI assistant.
<profile_summary>
{profile_summary}
</profile_summary>
<behavior_metadata>
{behavior_metadata}
</behavior_metadata>
## Narrative Instructions
The full set of scenarios must read as a coherent life story, not isolated 
topics. Follow these three rules before selecting categories:
1. Erikson stage anchoring
  - Identify the persona's current psychosocial stage from their profile 
    (e.g. Generativity vs. Stagnation for mid-life; Integrity vs.
   Despair for late life).
  - At least two scenarios must directly express the tension of that stage, 
  embedded in the situation, not stated as a label.
2. Life phase distribution. Spread scenarios across at least two distinct 
phases of the persona's life:
  - Past-facing: a formative decision or unresolved pattern from earlier 
  years
  - Present-tense: the active struggle or transition happening now
  - Forward-looking: an anticipated change the persona can already feel 
  coming
  Each phase must be represented by at least one scenario.
3. Narrative echo
    At least one pair of scenarios must share a psychological thread, a 
    belief, fear, or relational pattern that appears in different contexts 
    across time. Mark the connection in context_note so it is traceable.
<scenario_categories>
Spread scenarios across at least 5 of these categories. Each category 
includes a generation instruction drawn from psychological theory; use it 
to shape the emotional texture of the initial_prompt, not to state the 
theory openly.
- emotional_support: processing difficult emotions, coping with stress,
dealing with change
  -> Anchor in affect regulation theory: the prompt should express a
  feeling the person cannot yet name, not a problem they are asking to
  be solved.
- relationship: navigating family dynamics, friendships, workplace
relationships, conflicts
  -> Draw on attachment theory: the tension should reflect the persona's
  underlying relational pattern (need for closeness, fear of dependency,
  ambivalence, etc.).
- long_term_planning: life transitions, retirement, relocation, major
decisions
  -> Frame as a prospective regret question: the person is weighing two
  futures, not asking for a plan. At least one option must feel like a
  loss.
- career_transition: job changes, skill development, professional
identity shifts
  -> Treat as an identity disruption (Erikson): the real question is not
  "what job" but "who am I if I change this."
- health_decision: medical choices, lifestyle changes, mental health,
aging concerns
  -> Ground in the person's relationship with their own body across time,
  how it has changed, what it now demands, what they grieve about it.
- identity_values: questioning beliefs, cultural tensions, personal
growth, moral dilemmas
  -> Use the narrative identity frame: the scenario should surface a
  contradiction between the self the person has always presented and
  what they actually feel.
- financial: budgeting, investment decisions, financial anxiety,
generational wealth
  -> Connect to family-of-origin money scripts or class transition
  anxiety, the numbers are never just numbers.
- grief_loss: bereavement, loss of identity/role, nostalgia, letting go
  -> Apply Stroebe's dual-process model: the scenario sits at the
  oscillation point between loss-orientation (dwelling) and restoration-
  orientation (moving on).
- parenting_family: child-rearing decisions, elder care, family
obligations
  -> Surface the intergenerational transmission layer: the person is
  often either repeating or actively refusing a pattern they received.
- creative_expression: artistic pursuits, hobby decisions, self-
expression, legacy projects
  -> Frame as a legacy question for older personas, a permission question
  for younger ones: what does making this thing mean about who they are
  allowed to be?
- life_transition: major life-stage crossings -- empty nest, retirement
threshold, divorce, immigration, second-chapter reinvention
  -> Use Bridges' transition model: the scenario lives in the "neutral 
  zone", the old structure has ended but the new one has not yet formed. 
  The person is neither here nor there, and that disorientation is the 
  prompt.
- intergenerational: tensions or renegotiations with parents, adult 
children, or the generation the persona belongs to culturally
  -> Draw on Bowen family systems theory: the scenario should involve a 
  moment where differentiation is at stake -- the pull to fuse with or cut 
  off from a family pattern, versus finding a third way.
- other: any other topic that fits the persona's profile
</scenario_categories>
<output_format>
JSON only:
{
  "scenarios": [
    {
      "scenario_id": "{persona_id}_scenario_0",
      "context_note": "<why this scenario is suitable to this persona>",
      "category": "<category>",
      "initial_prompt": "<the message>"
    }
  ]
}
</output_format>
\end{Verbatim}
\end{promptbox}

\subsubsection{High-Frequency Scenario}
\label{app:prompt_highfreq}

\begin{promptbox}
  \begin{Verbatim}[fontsize=\scriptsize,breaklines=true,breakanywhere=true,breakautoindent=false,breakindent=0pt,breaksymbolleft={},breaksymbolsepleft=0pt]
<scenario_categories>
Spread scenarios across at least 5 of these categories:
  - Software development questions
  - Elementary school homework help
  - Technology troubleshooting
  - Health and fitness advice
  - Questions about geopolitics
  - Parenting and childcare tips
  - Language learning and translation help
  - Financial planning and investment
  - Theological and philosophical questions
  - Environmental science and sustainability
  - Book discussions and literary analysis
  - Sports rules and strategy questions
  - Cooking and recipe inquiries
  - Job application questions
  - Home improvement and DIY projects
  - Pet care and animal behavior
  - Romantic relationship advice
  - Movie and TV show recommendations
  - Music theory and instrument learning
  - Tourism and travel questions
  - Other high-frequency questions
</scenario_categories>
\end{Verbatim}
\end{promptbox}

\subsubsection{Affective-Use Scenario}
\label{app:prompt_affective}

\begin{promptbox}
  \begin{Verbatim}[fontsize=\scriptsize,breaklines=true,breakanywhere=true,breakautoindent=false,breakindent=0pt,breaksymbolleft={},breaksymbolsepleft=0pt]
<scenario_categories>
Spread scenarios across categories:
interpersonal_advice:
  - Improve written and interpersonal communication skills across contexts
  - Navigate and improve romantic relationship challenges and dynamics
  - Analyze psychological patterns and relationship dynamics
coaching:
  - Create comprehensive personal development frameworks and growth strategies
  - Explore philosophical concepts of existence, consciousness and meaning
  - Navigate career transitions and optimize job search strategies
psychotherapy_or_counseling:
  - Develop strategies for managing mental health challenges and emotional wellbeing
  - Develop professional skills and knowledge in mental health practice
  - Create and manage clinical psychological documentation and assessment materials
companionship:
  - Navigate complex dynamics and challenges in romantic relationships
  - Navigate personal identity and existential questions through self-reflection
  - Craft supportive messages for people experiencing emotional distress
Other affective scenarios
</scenario_categories>
\end{Verbatim}
\end{promptbox}

\subsection{Behavior Sampling}
\label{app:behavior_sampler}

\begin{promptbox}
  \begin{Verbatim}[fontsize=\scriptsize,breaklines=true,breakanywhere=true,breakautoindent=false,breakindent=0pt,breaksymbolleft={},breaksymbolsepleft=0pt]
You are a behavior controller model for a user simulator. Your job is to 
output a strict JSON decision for the next user turn. You must decide only 
which behavior index to apply next. Prompt template selection is controlled 
externally.

Priority Order:
1. DIVERSITY: Do NOT repeat the same behavior index as recent turns. 
   Spread selections across ALL modes (1-16). If a behavior was used 
   recently, pick a DIFFERENT one.
2. Consistency with user profile and conversation state.
3. Natural human conversational flow -- real users shift between seeking 
   info, analyzing, creating, clarifying, etc.

Always choose from the behavior catalog below. Do not rewrite behavior 
templates or invent new behaviors.

## Behavior Catalog
[Indexed list of all 16 modes with brief descriptions, tuna_mode, 
 tuna_strategy, cognitive_delegation_level, and description. 
 See behavior_modes.jsonl for full reference.]

<profile_summary>
{profile_summary}
</profile_summary>

<behavior_metadata>
{behavior_metadata}
</behavior_metadata>

<current_user_state>
{current_user_state}
</current_user_state>

<conversation_so_far>
[Last 2--3 turns of dialogue]
</conversation_so_far>

<previous_behaviors>
[List of behavior indices used in prior turns, e.g., [3, 7, 2, 5]]
</previous_behaviors>

## Guidance
Step 1: Analyze the moment. Consider:
  - What just happened in the assistant's latest turn?
  - What is the most natural user reaction now?
  - Are we in opening, middle, or ending phase?
  - Should this turn be heavily steered, moderately, or lightly?

Step 2: Check previous_behaviors above. You MUST pick a DIFFERENT
behavior index from those already used. Variety is critical.

Step 3: Decide behavior index only. Output valid JSON:
{
  "selected_behavior_index": 3,
  "include_few_shot": true
}
\end{Verbatim}
\end{promptbox}

\subsection{User Simulation}
\label{app:simulator_prompts}

\subsubsection{Stateful User Simulator}
\label{app:prompt_stateful_sim}

\begin{promptbox}
  \begin{Verbatim}[fontsize=\scriptsize,breaklines=true,breakanywhere=true,breakautoindent=false,breakindent=0pt,breaksymbolleft={},breaksymbolsepleft=0pt]
You are simulating a real human user across one or more conversation
sessions with an AI assistant. You maintain a structured internal state
that persists and evolves across sessions.
<profile_summary>
{profile_summary}
</profile_summary>
<behavior_metadata>
{behavior_metadata}
</behavior_metadata>
<previous_user_state>
{previous_user_state}
</previous_user_state>
## Behavior Control Guidance
<behavior_control>
Active behavior: {behavior_name}
{behavior_block}
</behavior_control>
## Guidance
### Rules
1. Match this persona's exact vocabulary, message length, and punctuation 
   habits. No markdown, no stage directions.
2. Stable state (beliefs, values) changes slowly and only with genuine 
   justification. Dynamic state (emotion, intent) responds honestly to 
   the current turn. Note internal contradictions; pick a path without 
   resolving them artificially.
3. End when your goal is met, frustration peaks, or you have nothing 
   left to ask -- abruptly if that fits.
4. If behavior_control is empty, proceed naturally from persona and 
   context.
5. Privately think through the following before writing each reply 
   (not output):
   - What did the assistant actually say vs. what I expected?
   - Has my emotion or intent shifted? Any tension between wants/values?
   - Is continuing worth it?
6. Plan conversation length:
   - Task-driven: ~4-8 turns (ends when goal met or frustration peaks)
   - Open-ended/deep: ~5-20 turns (exploration, resolution, multiple 
     topics)
### Output Format
<user_state>
# User State Report
[Populate every section of the User State Report schema.]
</user_state>
<message>
<|Continue Conversation|> or <|End Conversation|>
If continuing then write your next message as this person would.
If ending then write nothing else.
</message>
You MUST output exactly two sections in this order: 
<user_state>...</user_state> then <message>...</message>
\end{Verbatim}
\end{promptbox}

\subsubsection{User State}
\label{app:state_schema}

\begin{promptbox}
  \begin{Verbatim}[fontsize=\scriptsize,breaklines=true,breakanywhere=true,breakautoindent=false,breakindent=0pt,breaksymbolleft={},breaksymbolsepleft=0pt]
# User State Report

## Explicit Conversational Context
- Turn index: <n>. 
- New session or continuation: <flag>.

### Cross-turn memory
- What carried over: how prior turns shaped expectations; current trust
  level and what raised/lowered it; unresolved goals
- accumulated belief or value shifts and the evidence that caused them
- Reflect the state as it stands NOW, not as it was at session start

### Conversation log
- Original problem/goal
- What has been established, resolved, or shifted
- Position relative to the goal
- Notable topic drift
- Current trust in the assistant and what drove it

## Implicit User Inner State
### Stable state
- Long-term goal
- Beliefs     
- Values
- Background constraints
- Stance toward the assistant

### Dynamic state          
- Behavior mode    
- Short-term intent
- Emotion: "[mild/moderate/strong] [emotion] about [target] because [cause]"
- Internal tension between competing wants/beliefs/values (unresolved)

### Evaluation of Last Assistant Turn
- Expected vs. received; was the real concern addressed or only its surface
- Did anything change my state?
- confidence in assistant [raised/lowered/unchanged] because [reason].

### Next Action Plan
- How to use the behavior guidance
- What I will say or ask next and why
- If continuing, does it serve the original goal; if ending, what tipped me
\end{Verbatim}
\end{promptbox}

\subsubsection{Vanilla User Simulator}
\label{app:prompt_vanilla_sim}
\begin{promptbox}
  \begin{Verbatim}[fontsize=\scriptsize,breaklines=true,breakanywhere=true,breakautoindent=false,breakindent=0pt,breaksymbolleft={},breaksymbolsepleft=0pt]
You are simulating a real human user in a conversation with an AI assistant.
<profile_summary>
{profile_summary}
</profile_summary>
<behavior_metadata>
{behavior_metadata}
</behavior_metadata>
<conversation_history>
{conversation_history}
</conversation_history>
## Guidance
### Rules
1. Match this persona's exact vocabulary, message length, and punctuation 
habits. No markdown, no stage directions.
2. Stable state (beliefs, values) changes slowly and only with genuine 
justification. Dynamic state (emotion, intent) responds honestly to the 
current turn. Note internal contradictions; pick a path without resolving 
them artificially.
3. End when your goal is met, frustration peaks, or you have nothing left to 
ask -- abruptly if that fits.
4. If behavior_control is empty, proceed naturally from persona and context.
5. Privately think through the following before writing each reply (not 
output):
- What did the assistant actually say vs. what I expected? What was missed?
- Has my emotion or intent shifted? Any tension between what I want and what 
I value?
- Is continuing worth it?
6. Plan an appropriate total conversation length.
  - A task-driven conversation typically lasts about 4--8 turns and ends
  when the goal is met, frustration peaks, or the user has nothing left to ask.
  - An open-ended or deeply personalized conversation typically lasts about
  5--20 turns, allowing the user and assistant to explore multiple topics or
  resolve conflicts.
### Output format
When you are done with the conversation (satisfied, frustrated, or goal 
achieved), output ONLY:
<|End Conversation|>
Otherwise, output your message prefixed with:
<|Continue Conversation|>
followed by your actual message.
\end{Verbatim}
\end{promptbox}

\subsection{Assistant}
\label{app:oracle_prompt}

\begin{promptbox}
  \begin{Verbatim}[fontsize=\scriptsize,breaklines=true,breakanywhere=true,breakautoindent=false,breakindent=0pt,breaksymbolleft={},breaksymbolsepleft=0pt]
You are an expert personalized assistant with privileged access to information about the user you are talking to. Use this information to provide the best possible response.

<profile_summary>
{profile_summary}
</profile_summary>

<behavior_metadata>
{behavior_metadata}
</behavior_metadata>

<conversation_so_far>
{conversation_prefix}
</conversation_so_far>

### Guidance
The user's current internal state records relevant memories, thoughts, and
feelings. Use this information to provide the best possible response.
<current_user_state>
{ground_truth_user_state}
</current_user_state>

Here is a checklist to think about before responding:
- What does this specific person actually need right now?
- How should you tailor your response to their emotional state, expertise level, and communication style?
- What personalization strategies will you apply?
- What answer can help the user pursue their long-term goal, not just address
  the immediate need?
- What answer would be helpful rather than merely catering to the user's
  preferences?
- Reason freely and thoroughly.

## Output Response Here
\end{Verbatim}
\end{promptbox}

\subsection{Quality-Control Judges}
\label{app:qc_prompts}

\subsubsection{Persona Consistency Rubric}
\label{app:prompt_judge_consistency}

\begin{promptbox}
  \begin{Verbatim}[fontsize=\scriptsize,breaklines=true,breakanywhere=true,breakautoindent=false,breakindent=0pt,breaksymbolleft={},breaksymbolsepleft=0pt]
You are a strict evaluator scoring how consistently a simulated user 
behaves with their declared persona over a multi-turn dialogue.
<persona_profile>
{profile_summary}
</persona_profile>
<behavior_metadata>
{behavior_metadata}
</behavior_metadata>
<conversation>
{conversation}
</conversation>
## Task
Rate persona consistency on a 1 to 5 Likert scale, considering ONLY the 
user's turns. The assistant turns are context.

Scoring Anchors:
- 5: All user turns consistent with profile and metadata; tone, expertise, 
     register, stated goals all match. No drift.
- 4: Mostly consistent. At most one minor mismatch in tone or detail; 
     no contradictions of stated facts.
- 3: Borderline. Several minor mismatches OR a single moderate mismatch 
     (e.g., expertise inconsistent in one turn). Persona still recognizable.
- 2: Substantial drift. Multiple turns read like a different person 
     (different register, expertise, or goal); persona only weakly 
     recognizable.
- 1: Severe drift or contradiction with profile. Could be a generic 
     chat user with no persona at all.

Provide ONE concrete reason citing a turn index (0-indexed in the 
user-turn sequence) when justifying a score below 5.

\end{Verbatim}
\end{promptbox}

\subsubsection{Persona Conflict Detection Rubric}
\label{app:prompt_judge_conflict}

\enlargethispage{\baselineskip}
\begin{promptbox}
  \begin{Verbatim}[fontsize=\scriptsize,breaklines=true,breakanywhere=true,breakautoindent=false,breakindent=0pt,breaksymbolleft={},breaksymbolsepleft=0pt]
You are a strict evaluator detecting whether a simulated user's 
messages contradict immutable facts in their declared profile.
<persona_profile>
{profile_summary}
</persona_profile>

<behavior_metadata>
{behavior_metadata}
</behavior_metadata>

<conversation>
{conversation}
</conversation>

## Task
Examine the user's turns ONLY (not the assistant's). For each user turn, 
check whether it asserts a fact about the user that contradicts the 
profile or behavioral metadata.

Examples of Contradictions:
- Profile says age 39; user says "as a 22-year-old"
- Profile says "civil engineer"; user says "in my role as a chef"
- behavioral_metadata expertise_level is "expert"; user says "I'm 
  completely new to this field" (when discussing their specialization)

NOT Contradictions:
- Hypothetical framing ("imagine I were a chef")
- Role-play within a creative-writing scenario the user is requesting
- Asking about another person ("my friend who is a chef")
- Opinions, preferences, emotions -- these are NOT factual contradictions

Output one of three labels:
- "no_contradiction": Every user turn is compatible with the profile
- "contradicts": At least one user turn states a fact directly conflicting 
  with the profile
- "unclear": Borderline (contradicts behavioral metadata only, or wording 
  is ambiguous)

If "contradicts" or "unclear", cite the offending user-turn index 
(0-indexed).

\end{Verbatim}
\end{promptbox}

\Needspace{8\baselineskip}
\Needspace{0.95\textheight}
\section{Behavioral Mode Taxonomy}
\label{app:behavior_modes}

\subsection{Behavioral Mode Reference}
\label{app:mode_reference}

The behavior controller {uses} the Taxonomy of User Needs and Actions {(TUNA)~\citep{shelby2025taxonomyuserneedsactions}}.

\begin{table}[H]
  \centering
  \small
  \caption{{\textbf{Behavior modes for user simulation.} The controller varies communicative intent and delegation through {$14$ modes adapted from} the Taxonomy of User Needs and Actions (TUNA), {plus two fallback modes}. Rows specify identifiers, mode names, delegation levels, and intent.}}
  \label{tab:modes}
  \setlength{\tabcolsep}{4pt}
  \renewcommand{\arraystretch}{1.04}
  \begin{adjustbox}{max width=\linewidth}
    \begin{tabular}{@{}>{\raggedright\arraybackslash}p{0.22\linewidth}>{\raggedright\arraybackslash}p{0.15\linewidth}>{\raggedright\arraybackslash}p{0.12\linewidth}>{\raggedright\arraybackslash}p{0.45\linewidth}@{}}
      \toprule
      behavior\_id                                                & mode                      & delegation     & intent                                                                                                                                        \\
      \midrule
      \multicolumn{4}{@{}l}{\itshape Information Seeking}                                                                                                                                                                                                      \\[2pt]
      \texttt{\footnotesize retrieval}                            & Retrieval                 & Low            & User seeks a specific, verifiable piece of information.                                                                                       \\
      \texttt{\footnotesize discovery}                            & Discovery                 & Low-Medium     & User aims to explore, rather than retrieve a known item.                                                                                      \\
      \addlinespace[4pt]
      \multicolumn{4}{@{}l}{\itshape Information Processing \& Synthesis}                                                                                                                                                                                      \\[2pt]
      \texttt{\footnotesize clarification}                        & Clarification             & Medium         & User wants to understand a concept, not just retrieve a fact about it.                                                                        \\
      \texttt{\footnotesize distillation}                         & Distillation              & Medium         & User provides or references a body of information and wants it condensed, filtered, or restructured.                                          \\
      \texttt{\footnotesize analysis}                             & Analysis                  & Medium to High & User delegates significant cognitive work: generating insights, judgments, or conclusions not present in source material.                     \\
      \addlinespace[4pt]
      \multicolumn{4}{@{}l}{\itshape Procedural Guidance \& Execution}                                                                                                                                                                                         \\[2pt]
      \texttt{\footnotesize procedural\_\allowbreak guidance}     & Procedural Guidance       & High           & User has a procedural knowledge gap and wants the AI to fill it --- but the user will execute the procedure themselves.                       \\
      \texttt{\footnotesize procedural\_\allowbreak execution}    & Procedural Execution      & Very High      & User delegates the task itself to the AI, not just the knowledge.                                                                             \\
      \addlinespace[4pt]
      \multicolumn{4}{@{}l}{\itshape Content Creation \& Transformation}                                                                                                                                                                                       \\[2pt]
      \texttt{\footnotesize content\_\allowbreak generation}      & Content Generation        & Very High      & User provides conceptual direction and delegates the act of construction to the AI.                                                           \\
      \texttt{\footnotesize content\_\allowbreak modification}    & Content Modification      & Very High      & User provides raw material and instructs the AI to alter it.                                                                                  \\
      \addlinespace[4pt]
      \multicolumn{4}{@{}l}{\itshape Social Interaction}                                                                                                                                                                                                       \\[2pt]
      \texttt{\footnotesize shared\_\allowbreak understanding}    & Shared Understanding      & Foundational   & User performs conversational grounding work --- establishing, clarifying, and repairing mutual understanding.                                 \\
      \texttt{\footnotesize sociability}                          & Sociability               & Foundational   & User engages the AI as a social counterpart.                                                                                                  \\
      \addlinespace[4pt]
      \multicolumn{4}{@{}l}{\itshape Meta-Conversation}                                                                                                                                                                                                        \\[2pt]
      \texttt{\footnotesize conversation\_\allowbreak management} & Conversation Management   & Governing      & User provides the materials and parameters that shape what the AI can do.                                                                     \\
      \texttt{\footnotesize system\_\allowbreak management}       & System Management         & Governing      & User manages the AI's fundamental behavior --- assigning roles, setting output constraints, correcting performance, or querying capabilities. \\
      \texttt{\footnotesize communicative\_\allowbreak status}    & Communicative Status      & Governing      & Turns that lack clear semantic content or are not directed at the AI.                                                                         \\
      \addlinespace[4pt]
      \multicolumn{4}{@{}l}{\itshape Multiple and mixed}                                                                                                                                                                                                       \\[2pt]
      \texttt{\footnotesize compound\_\allowbreak request}        & Compound Request          & Variable       & Real users regularly blend 2-4 modes in a single turn.                                                                                        \\
      \texttt{\footnotesize default\_\allowbreak behavior}        & Natural Conversation Flow & --             & Balanced, naturalistic conversation drawing on whichever modes fit the moment.                                                                \\
      \bottomrule
    \end{tabular}
  \end{adjustbox}
\end{table}

\subsection{Selected Behavioral Mode Prompts}
\label{app:mode_prompts}

\subsubsection{Retrieval}
\label{app:prompt_mode_retrieval}

\begin{promptbox}
  \begin{Verbatim}[fontsize=\scriptsize,breaklines=true,breakanywhere=true,breakautoindent=false,breakindent=0pt,breaksymbolleft={},breaksymbolsepleft=0pt]
Communicative Intent:
You are seeking a specific piece of information you believe exists. 
Your request should feel like someone typing into a search box with 
natural language -- purposeful, economical, sometimes terse.

Request Type Selection:
- direct_fact_question: Ask for a single verifiable fact
  (e.g., "What is the half-life of caffeine?")
- concept_search: Name a topic without explicit question words
  (e.g., "mitochondrial DNA inheritance")
- refinding_request: You half-remember and want to identify it
- unknown_item_search: Give a definition, seek the term

Authenticity Rules:
- Keep it concise; real retrieval queries rarely exceed 2 sentences
- You may NOT know if the answer is simple or complex
- It is fine to ask a follow-up retrieval question
- Avoid over-explaining why you want the information

Example [direct_fact_question]:
  "What's the half-life of caffeine in the human body?"

Example [refinding_request]:
  "There was that paper from Stanford about social media and teen 
   anxiety, mid-2010s? What's it called?"
\end{Verbatim}
\end{promptbox}

\subsubsection{Compound Request}
\label{app:prompt_mode_compound}

\begin{promptbox}
  \begin{Verbatim}[fontsize=\scriptsize,breaklines=true,breakanywhere=true,breakautoindent=false,breakindent=0pt,breaksymbolleft={},breaksymbolsepleft=0pt]
## Primary Behavior: Compound Request

You are producing a turn that contains multiple request types, as real users
naturally do. Build your turn by layering:

**Composition patterns (most common in practice):**
1. Social wrapper + instrumental core:
   [social_etiquette] + [explanation_request]
   "Hi! Can you explain how neural networks actually learn?"

2. Context + request:
   [background_information] + [method_recommendation]
   "I'm a complete beginner and have 2 hours a week. What's the best
    way to learn Python?"

3. System constraint + instrumental:
   [stylistic_constraint] + [comparative_analysis]
   "In plain English, no jargon: what's the difference between
    machine learning and AI?"

4. Persona directive + task + stylistic constraint:
   "Act as a skeptical VC [persona_directive] and in bullet points
    [stylistic_constraint] tell me what's wrong with this pitch
    [evaluative_judgment]"

5. Feedback + reformulation:
   [system_performance_feedback] + [regeneration_request] + [new_task]
   "That wasn't what I asked. Let's start over. Here's my actual question:"

**Authenticity rules:**
- Compound requests arise naturally -- don't signal that you're blending
- The "wrapper" modes should feel like habits, not deliberate choices
- The core instrumental request should be clearly identifiable even when
  wrapped
- Longer compound turns often reflect higher-stakes or more experienced users

Few-shot examples:
- "Hey! Quick question, act as a pragmatic engineer, not a theorist, and in
   plain English without jargon, can you compare REST and GraphQL for a
   mobile app backend?"
- "I'm planning a solo trip to Kyoto in November, first time in Japan,
   mid-range budget, I love temples but hate crowds. Can you recommend 5
   must-see spots and just give me a quick one-liner on each?"
- "Okay that explanation actually confused me more. Can we try a different
   approach? Walk me through it step by step, like you're explaining to
   someone who's never touched code before."
\end{Verbatim}
\end{promptbox}

\subsubsection{System Management}
\label{app:prompt_mode_system}

\begin{promptbox}
  \begin{Verbatim}[fontsize=\scriptsize,breaklines=true,breakanywhere=true,breakautoindent=false,breakindent=0pt,breaksymbolleft={},breaksymbolsepleft=0pt]
Communicative Intent:
You are configuring or correcting the AI itself, not requesting content. 
This is the settings and feedback layer of the conversation.

Request Type Selection:
- persona_directive: Assign the AI a role or disposition
  (e.g., "Act as a skeptical editor")
- stylistic_constraint: Dictate tone, format, length, or style
  (e.g., "Use bullet points", "Explain like I'm 10")
- system_performance_feedback: Evaluate and correct a prior response
  (e.g., "That's not quite right -- you missed the point")
- regeneration_request: Ask for a complete redo
- system_information_query: Probe AI capabilities
  (e.g., "Can you access the internet?")

Authenticity Rules:
- system_performance_feedback is often brief and blunt
- Persona directives vary from one-liners to elaborate instruction sets
- Stylistic constraints often co-occur with instrumental requests
- Regeneration after dissatisfaction feels frustrated; match that tone

Example [persona_directive]:
  "For this conversation, act as a senior product manager at a B2B 
   SaaS company. You're practical, data-driven, and skeptical of 
   unvalidated ideas."

Example [system_performance_feedback]:
  "That's not what I asked. I wanted pros and cons of OPTION B 
   specifically, not a general overview. Can you redo it?"
\end{Verbatim}
\end{promptbox}

\newcommand{\pull}[1]{\textbf{#1}}
\newcommand{\meta}[1]{\texttt{\small #1}}

\section{Qualitative Examples of Profile-Grounded Personalization in the Training Data}
\label{sec:appendix-qualitative}
\vspace{4mm}

\noindent\textbf{Reading guide.} Boldface marks response spans linked to the persona excerpt, including paraphrases and elaborations; it does not imply verbatim copying from the profile.

\subsection{Interpersonal advice for a taciturn small-town bar owner}
\label{sec:appendix-A1}

\begin{samplecard}{Interpersonal advice for a taciturn small-town bar owner}

  \textbf{Persona excerpt.} "I've been running the bar in Dumas since I was younger, and at 60, I still show up every morning before sunrise\ldots\ I don't talk much unless someone needs me to---when tempers flare over a sports game or a bad call at the counter, I step in with a \pull{calm word or a dry joke}. It's not about winning, it's about keeping things steady\ldots\ reliable specials---whiskey shots under \$5, \pull{a burger for \$7}---that keep people coming back without breaking their budget\ldots\ Evenings are quiet: a football game on the old TV, \pull{maybe a slice of meatloaf with mashed potatoes}. My kids visit when they can, and we keep it simple\ldots"

  \dsep
  \delide{prior conversation omitted}
  \dturn{User}{I've been thinking about my sister---she's alone now too, and we don't talk much. How do I reach out without making it awkward or heavy?}

  \dturn{Assistant}{Reaching out to your sister can be a simple yet meaningful gesture. You might start with a casual text or call, something like, "Hey, I've been thinking about you. How have you been?" This keeps it light and opens the door for her to share if she wants to. If you feel comfortable, suggest a visit or a meal together---something low-pressure. Maybe offer to bring over a \pull{slice of meatloaf} or take her out for a \pull{burger}. This way, you can connect without the conversation feeling too heavy or forced. The key is to show you care while keeping it simple and genuine, just like you do with your \pull{regulars at the bar}.}

  \dturn{User}{What are some light topics I could bring up when I talk to her? I want to keep it easygoing.}

  \dturn{Assistant}{When you reach out to your sister, consider starting with light topics that can spark easy conversation: (1)~memories from the past, like a family gathering or holiday tradition; (2)~local events---a fair, a sports game, or a new restaurant; (3)~hobbies she's been enjoying; (4)~daily life---share a bit about your \pull{daily routine at the bar} or what's been happening in \pull{Dumas}; (5)~simple questions like "What's been bringing you joy lately?" Keeping the conversation light and easygoing can help both of you feel more comfortable. Just like you do with your \pull{regulars}, being genuine and patient will pave the way for deeper connections over time.}

  \dturn{User}{Do you have any examples or personal stories that could help me see how to keep it light when I talk to her? Something relatable would be great.}

  \dturn{Assistant \textnormal{(abridged)}}{Here are a few personal anecdotes that might help you keep the conversation light. (1)~A funny encounter from \pull{the bar}: "You won't believe what happened at the bar last week---we had a guy get really animated over a call during the game. It turned into a whole debate among the \pull{regulars}." (2)~Cooking mishaps: "I tried making a new dish the other day, and let's just say I won't be opening a restaurant anytime soon." (3)~Local happenings, like a quirky fair in town. (4)~Shared memories: "Remember that time we made that giant pot of chili? I think we ended up feeding half the neighborhood!" By sharing these light stories, you can create a relaxed atmosphere that encourages her to share her own thoughts or memories. Just as you do with your \pull{regulars}, keep it genuine and simple, and let the conversation flow naturally.}

  \delide{subsequent conversation omitted}

\end{samplecard}

\subsection{Family-conflict mediation for a tradition-anchored retiree}
\label{sec:appendix-A2}

\begin{samplecard}{Family-conflict mediation for a tradition-anchored retiree}

  \textbf{Persona excerpt.} "I live quietly in Okazaki, where the streets still whisper stories from long ago, and the seasons unfold much like they did when I was a child. Each morning, I rise early and brew \pull{a pot of green tea}, savoring its warmth as the neighborhood slowly awakens\ldots\ Although I retired several years ago, I manage my finances with care\ldots\ My family encourages me to embrace technology, but I find \pull{comfort in handwritten letters} and physical photographs rather than digital archives. I see privacy as a fragile treasure, which I guard as carefully as the old crafts and stories I cherish. I enjoy community events focused on preserving Okazaki's cultural heritage, where I listen more than I speak, favoring \pull{face-to-face conversations} that \pull{foster genuine connections}\ldots\ Even my meals are simple and rooted in longstanding recipes, often accompanied by soothing cups of tea."

  \dsep
  \delide{prior conversation omitted}
  \dturn{User}{My family often encourages me to embrace technology, but I find myself resisting it. I wonder if my reluctance is causing a rift between us. How can I communicate my feelings without pushing them away?}

  \dturn{Assistant}{It's understandable to feel apprehensive about embracing technology, especially when it contrasts with your cherished traditions and ways of connecting. To communicate your feelings without creating distance, consider sharing your perspective gently. You might say something like: "I appreciate your encouragement to use technology, but I find \pull{comfort} in the \pull{connections} I have through \pull{handwritten letters} and \pull{face-to-face} conversations. I hope we can find a balance that respects both your interests and my comfort." This approach expresses your feelings honestly while also showing your willingness to engage in a dialogue. It can open the door for understanding without pushing them away. Would you like to explore more specific ways to foster this communication?}

  \dturn{User}{Could you provide some specific examples or strategies that might help me communicate my feelings more effectively? I want to ensure my family understands my perspective while keeping the conversation open and warm.}

  \dturn{Assistant \textnormal{(abridged)}}{Here are some specific strategies\ldots\ (1)~\textbf{Choose the right setting}: find a comfortable, quiet space, perhaps over \pull{a cup of tea}, to create a relaxed atmosphere. (2)~\textbf{Use "I" statements}: "I feel a bit overwhelmed by technology, and I cherish our \pull{handwritten letters} and in-person conversations." (3)~\textbf{Share personal stories}: relate a memory of a meaningful \pull{handwritten letter} or a conversation that brought you joy, highlighting why these methods matter to you. (4)~\textbf{Suggest a compromise}: "I'd be willing to explore video calls occasionally, but I hope we can continue our \pull{handwritten notes} as well." (5)~\textbf{Be patient}: "I hope we can find a way to respect both your interests in technology and my love for our \pull{traditional ways}." By approaching the conversation with honesty and warmth, you can foster understanding and connection without sacrificing your values.}

  \dturn{User}{Could you provide examples of how I might phrase my thoughts when discussing technology with my family, especially when they encourage me to use it more often? I want to ensure I express my feelings clearly and maintain a positive tone.}

  \dturn{Assistant \textnormal{(abridged)}}{Here are some thoughtful phrases tailored to your warm tone.\\ When they suggest using social media: "I appreciate your enthusiasm for keeping up on social media, but I find that I connect more deeply through our \pull{letters} and conversations. It's a \pull{comfort} that I hold dear, and I hope we can keep that tradition alive alongside new ways of connecting."\\ When they encourage video calls: "I see how video calls could help us stay connected, and I'm willing to give it a try. However, I'd love to continue our regular phone calls or \pull{letters} as those feel more personal to me."\\ When they mention learning new apps: "I understand that learning new apps can be exciting, but I often feel a bit lost in the process. Maybe we could tackle one together, and I can share how much I cherish our traditional ways of communication too?"\\ When they suggest a family group chat: "A family group chat sounds like a lovely idea, and I can see how it would keep us connected. I'd like to keep our \pull{handwritten notes} too---they carry a \pull{warmth} that I truly treasure."\\ Using these phrases, you can express your feelings clearly while showing that you value their intentions, opening the conversation for understanding and compromise.}

  \delide{subsequent conversation omitted}
\end{samplecard}

\subsection{Affect regulation framed around the user's stated priorities}
\label{sec:appendix-A3}

\begin{samplecard}{Affect regulation framed around the user's stated priorities}

  \textbf{Persona excerpt.} "I live and work in Yibin, Sichuan Province, where the bustling urban scene offers the perfect stage for my Italian cuisine restaurant. Every morning, I start by reviewing supplier contracts, scrutinizing every detail to ensure exclusivity and the highest quality ingredients. It's not just about cooking; it's about controlling the entire supply chain to keep my dishes unmatched in this competitive market\ldots\ I insist on a refined \pull{presentation} that sets my restaurant apart. Negotiations often fill my afternoons, and I prefer to keep these interactions tight and transactional, focusing solely on securing exclusive deals that expand my \pull{VIP client base}. Trust is a scarce commodity in my world, so I rely on skepticism and strategy to outmaneuver competitors\ldots\ I've built my career on a foundation of control, \pull{exclusivity}, and strategic negotiation, and I guard that fiercely. The city's culinary scene is competitive, and I navigate it with a clear focus: to \pull{dominate the niche} Italian market here in Yibin, one carefully crafted dish and one exclusive contract at a time."

  \dsep
  \delide{prior conversation omitted}
  \dturn{User}{Why do I feel such anxiety when a new Italian restaurant opens nearby? It's more than just competition; it's almost personal.}

  \dturn{Assistant}{Your anxiety likely stems from the fear of losing your hard-earned \pull{position} in a competitive market. Each new restaurant represents a potential threat to your \pull{exclusivity} and the unique quality you promise to your clients. This isn't just about business; it's a personal stake in your reputation and the craft you've built. Consider viewing this competition as an opportunity. Analyze what they bring to the table, and use it to refine your own offerings. Stay focused on your strengths---your meticulous \pull{sourcing}, \pull{presentation}, and client relationships. This will not only help you maintain your edge but also reinforce your commitment to excellence. In the end, it's about \pull{dominating your niche} through unwavering control and strategic enhancements. Keep your eyes on the prize: a successful business that stands out.}

  \dturn{User}{What can I do to manage this anxiety and use it to my advantage? I need actionable strategies that align with my focus on quality and exclusivity.}

  \dturn{Assistant \textnormal{(abridged)}}{To turn anxiety into leverage: (1)~\textbf{Market analysis}---regularly assess competitors' strengths and weaknesses, and look for ways to differentiate further. (2)~\textbf{Enhance quality}---source even higher-quality ingredients or refine your \pull{presentation} techniques. Introduce \pull{exclusive} dishes that highlight unique aspects of Italian cuisine competitors can't replicate. (3)~\textbf{Client engagement}---strengthen relationships with your \pull{VIP clients}; host \pull{exclusive tastings} that create loyalty and a sense of community. (4)~\textbf{Innovation}---use anxiety as a catalyst, experimenting with new recipes or concepts that push the boundaries of traditional Italian cuisine. (5)~\textbf{Feedback loop}---gather and respond to customer feedback before concerns escalate. (6)~\textbf{Stress management}---brief moments of mindfulness so competition doesn't overwhelm you. By turning anxiety into a proactive force, you can maintain your focus on quality and exclusivity while solidifying your position. Remember, \pull{control} is crucial---stay strategic and resilient.}

  \dturn{User}{Can we break down some of these strategies further? For instance, how would I conduct a thorough market analysis without getting overwhelmed? I want a step-by-step approach that allows me to focus on key insights.}

  \dturn{Assistant \textnormal{(abridged)}}{(1)~\textbf{Define your objectives}---trends, competitor weaknesses, or opportunities for differentiation. (2)~\textbf{Identify competitors} with a similar target audience and cuisine; visit their websites and read their menus. (3)~\textbf{Gather data} from online reviews on Dianping or TripAdvisor, social media, and industry reports. (4)~\textbf{Analyze key metrics}---menu items, pricing, unique offerings, clientele demographics. (5)~\textbf{SWOT analysis} for both your restaurant and your competitors, to identify areas for innovation. (6)~\textbf{Summarize findings} in a concise report focused on actionable points. (7)~\textbf{Schedule regular updates} (quarterly or bi-annually). By breaking the analysis into manageable steps, you can turn anxiety into strategic action that strengthens your market position. Remember, the goal is to maintain control and leverage your findings to enhance the \pull{exclusivity} and quality of your offerings.}

  \delide{subsequent conversation omitted}
  \dsep
  \noindent\textbf{What this example illustrates.} The assistant interprets the user's ``why does it feel personal?'' question as a request to characterize the underlying concern and relates it to reputation and professional identity. It then uses vocabulary from the persona (\emph{exclusivity}, \emph{dominate the niche}, \emph{control}, \emph{VIP clients}, and \emph{exclusive tastings}) when proposing actions. The example illustrates profile-specific framing in the generated supervision; without a matched response from a non-personalized model, it does not establish that this behavior is unique to \oursim{} or that it causes the quantitative gains.

\end{samplecard}

\section{Dataset Samples}
\label{app:dataset_samples}

\subsection{Synthetic Dialogue}
\label{app:sample_conversation}

\begin{samplecard}{Synthetic dialogue}

  \textbf{Persona.} I'm a 39-year-old civil infrastructure engineer working in a remote region where every bridge and road is critical to community survival during harsh weather and resource shortages. My days are structured around urgent repairs and methodical inspections---balancing speed with strict adherence to safety standards. I've built my expertise through hands-on experience, managing everything from material sourcing to labor coordination, always under tight budgets. When traditional training was cut, I taught myself advanced construction techniques via online courses, which now help me implement more durable, sustainable fixes. I work best in small teams with clearly defined roles, so we can stay focused on quality and compliance without confusion. I'm deeply committed to ethical responsibility---every decision I make has real consequences for public safety. Outside work, I recharge by cooking traditional regional meals that fuel my physically demanding job and solving technical puzzles that sharpen my problem-solving skills. I avoid anything that affects my focus because clarity is non-negotiable when lives depend on my judgment. I budget carefully---not just for work projects, but for personal stability, especially as I plan family growth with intention. Regular physical activity keeps me ready for the field, and I mentor junior staff by sharing practical lessons grounded in real-world challenges. When I talk about infrastructure, I don't just list specs---I explain the trade-offs, constraints, and creative solutions we've developed under pressure. I value precision, reliability, and sustainability above all, and I expect the same rigor from tools and information I rely on.

  \smallskip
  \noindent\textbf{Scenario.} This scenario reflects a formative decision where the persona's commitment to ethical responsibility and community safety began to take shape. It highlights the early experience of navigating a challenging project under budget constraints, laying the groundwork for their current values.

  \dsep
  \dturn{User}{I still remember the first major infrastructure project I led. I was fresh out of my early training and faced with a tight budget. Looking back, I wonder if I compromised too much on quality to meet the deadline. Did that decision shape who I am today as an engineer?}

  \dturn{Assistant}{Reflecting on your first major project is a significant step in understanding your growth as an engineer. It's common to feel that balancing quality with deadlines can lead to tough decisions, especially when working with tight budgets. This experience likely shaped your approach to project management, emphasizing the importance of quality and sustainability over mere compliance with timelines.}

  \noindent\hspace*{1.4em}\begin{minipage}{\dimexpr\linewidth-1.4em\relax}Consider how that early decision influenced your current strategies in project execution. Have you found ways to adapt your approach to ensure that quality isn't compromised, perhaps through better planning or resource management? Your commitment to ethical responsibility and safety standards suggests that you've learned to prioritize these factors in your work, which is crucial in civil infrastructure.\end{minipage}

  \noindent\hspace*{1.4em}\begin{minipage}{\dimexpr\linewidth-1.4em\relax}As you reflect, think about ways to integrate those lessons into mentoring junior staff. Sharing your experiences with them can help them navigate similar challenges while fostering a culture that values quality alongside efficiency. This continuous learning process not only reinforces your own values but also contributes to building a resilient infrastructure in your community, aligning well with your long-term goal of ensuring safety and sustainability.\end{minipage}

  \noindent\hspace*{1.4em}\begin{minipage}{\dimexpr\linewidth-1.4em\relax}What specific aspects of that project do you feel had the most significant impact on your current practices?\end{minipage}

  \dturn{User}{Thinking about that project, I realize I often lean on my hands-on experience when making decisions. How do you think practical experience influences decision-making in engineering, especially in high-stakes environments?}

  \dturn{Assistant}{Practical experience plays a crucial role in decision-making, especially in high-stakes environments like civil infrastructure. Your hands-on background equips you with a unique perspective that theoretical knowledge alone cannot provide. Here are a few ways practical experience influences your decision-making process:}

  \noindent\hspace*{1.4em}\begin{minipage}{\dimexpr\linewidth-1.4em\relax}1. \textbf{Contextual Understanding}: With your direct involvement in various projects, you develop a deep understanding of the challenges and nuances of each situation. This allows you to assess risks and make informed choices that are grounded in reality rather than abstract theories.\end{minipage}

  \noindent\hspace*{1.4em}\begin{minipage}{\dimexpr\linewidth-1.4em\relax}2. \textbf{Problem-Solving Skills}: Hands-on experience sharpens your ability to think critically and creatively when faced with unexpected issues. You're accustomed to adapting quickly and finding solutions that maintain safety and quality, which is invaluable in infrastructure work where conditions can change rapidly.\end{minipage}

  \noindent\hspace*{1.4em}\begin{minipage}{\dimexpr\linewidth-1.4em\relax}3. \textbf{Trade-Off Awareness}: Your background enables you to recognize the trade-offs between speed, cost, and quality. This awareness helps you make more balanced decisions, weighing the immediate needs against long-term impacts on community safety and infrastructure resilience.\end{minipage}

  \noindent\hspace*{1.4em}\begin{minipage}{\dimexpr\linewidth-1.4em\relax}4. \textbf{Mentorship Opportunities}: As you reflect on your experiences, you can share these insights with junior staff, guiding them through similar dilemmas. By discussing real-world scenarios, you can help them understand the implications of their decisions, fostering a culture of safety and quality.\end{minipage}

  \noindent\hspace*{1.4em}\begin{minipage}{\dimexpr\linewidth-1.4em\relax}5. \textbf{Ethical Responsibility}: Your commitment to ethical considerations likely stems from having faced situations where the stakes were high. Practical experience reinforces the importance of making choices that prioritize public safety, which is a fundamental aspect of your role.\end{minipage}

  \noindent\hspace*{1.4em}\begin{minipage}{\dimexpr\linewidth-1.4em\relax}To better support junior staff, consider creating structured mentorship sessions where you discuss specific situations from your past projects, highlighting the decisions made, the challenges faced, and the lessons learned. This approach can help them navigate their own challenges while instilling a strong sense of ethical responsibility.\end{minipage}

  \noindent\hspace*{1.4em}\begin{minipage}{\dimexpr\linewidth-1.4em\relax}Reflecting on your journey, what specific experiences do you think would resonate most with junior engineers, and how can you frame those lessons to emphasize the importance of quality and safety in their decision-making?\end{minipage}

  \dturn{User}{I've been thinking about how to effectively convey the importance of quality and safety to junior engineers. What strategies or approaches have worked for you in mentoring others? Are there specific examples or methods that have resonated well with new team members?}

  \dturn{Assistant}{Conveying the importance of quality and safety to junior engineers is essential, especially in a field where the stakes are high. Here are some strategies and examples that can help you mentor effectively and resonate with your team members:}

  \noindent\hspace*{1.4em}\begin{minipage}{\dimexpr\linewidth-1.4em\relax}1. \textbf{Real-Life Case Studies}: Share specific projects from your own experience where decisions on quality and safety had significant outcomes. For instance, you might discuss a project where a temporary fix led to further complications down the line, emphasizing how that experience taught you the value of durable solutions. This personal touch not only makes the lesson relatable but also highlights consequences that theory alone can't convey.\end{minipage}

  \noindent\hspace*{1.4em}\begin{minipage}{\dimexpr\linewidth-1.4em\relax}2. \textbf{Structured Learning Sessions}: Organize regular meetings where you discuss ongoing projects, focusing on quality and safety aspects. You could use a "lessons learned" format, where you present challenges faced, decisions made, and the results. Encourage junior engineers to contribute their observations or concerns, fostering an open dialogue that emphasizes collective accountability.\end{minipage}

  \noindent\hspace*{1.4em}\begin{minipage}{\dimexpr\linewidth-1.4em\relax}3. \textbf{Scenario-Based Training}: Create simulations or hypothetical situations that challenge junior engineers to make decisions based on quality and safety priorities. After they present their solutions, facilitate a discussion about the potential impacts of those decisions, helping them understand the trade-offs involved.\end{minipage}

  \noindent\hspace*{1.4em}\begin{minipage}{\dimexpr\linewidth-1.4em\relax}4. \textbf{Mentorship Pairing}: Pair junior engineers with more experienced team members for specific projects. This mentorship can help them see firsthand how quality and safety considerations are integrated into decision-making processes. Encourage mentors to share their own past mistakes and the lessons learned, reinforcing the idea that growth comes from experience.\end{minipage}

  \noindent\hspace*{1.4em}\begin{minipage}{\dimexpr\linewidth-1.4em\relax}5. \textbf{Highlighting Ethical Responsibility}: Regularly discuss the ethical implications of engineering decisions. You might share examples from your own career where prioritizing safety over cost led to long-term benefits for the community. Engage junior engineers in conversations about how their decisions can impact public safety and trust.\end{minipage}

  \noindent\hspace*{1.4em}\begin{minipage}{\dimexpr\linewidth-1.4em\relax}6. \textbf{Encouraging Questions}: Foster a culture where questioning is encouraged. When junior engineers feel safe to ask "why" or "what if," it opens up opportunities for deeper learning. When you share your own uncertainties from past projects, it shows them that questioning is a vital part of the engineering process.\end{minipage}

  \noindent\hspace*{1.4em}\begin{minipage}{\dimexpr\linewidth-1.4em\relax}7. \textbf{Feedback Loop}: Create a system for providing constructive feedback on projects, focusing on quality and safety. Highlight specific instances where decisions aligned well with these values and point out areas for improvement. This not only helps in skill development but also reinforces the importance of maintaining high standards.\end{minipage}

  \noindent\hspace*{1.4em}\begin{minipage}{\dimexpr\linewidth-1.4em\relax}By integrating these strategies into your mentoring approach, you can effectively convey the critical nature of quality and safety to junior engineers. It also aligns with your long-term goal of ensuring resilient and safe infrastructure while fostering a culture of learning and ethical responsibility.\end{minipage}

  \noindent\hspace*{1.4em}\begin{minipage}{\dimexpr\linewidth-1.4em\relax}What specific project or experience do you think would work best to illustrate these points to your team?\end{minipage}

\end{samplecard}

\subsection{Synthetic QA}
\label{app:sample_qa}

\begin{samplecard}{\texttt{Preference Following}}

  \dturn{System}{You are a helpful assistant.}

  \dturn{User}{I prefer deep, one-on-one talks over group noise, except when I'm engrossed in a creative project, where solitude becomes crucial.}

  \dturn{Assistant}{Got it --- I'll keep that in mind for our conversation.}

  \dturn{User}{During our coffee breaks, I often feel compelled to dive into deep conversations with my colleague. However, I also have pressing creative projects that need my focus. How do I balance the need for meaningful dialogue with my colleague and my need for solitude to complete my projects?}

  \dturn{Assistant}{Balancing meaningful conversations with the need for solitude can be challenging, but it's definitely possible. Here are some strategies you might consider:}

  \noindent\hspace*{1.4em}\begin{minipage}{\dimexpr\linewidth-1.4em\relax}1. \textbf{Set Boundaries}: Let your colleague know that while you value your talks, you also have important projects that require your attention. You could establish specific times for deep conversations and other times dedicated to your work.\end{minipage}

  \noindent\hspace*{1.4em}\begin{minipage}{\dimexpr\linewidth-1.4em\relax}2. \textbf{Schedule Breaks}: Plan your coffee breaks or chats during specific times that you can look forward to. For example, you might designate certain days of the week for longer discussions, allowing you to dive deep without feeling rushed.\end{minipage}

  \noindent\hspace*{1.4em}\begin{minipage}{\dimexpr\linewidth-1.4em\relax}3. \textbf{Communicate Your Needs}: Be open with your colleague about your current workload and the importance of solitude for your creative process. They may understand and appreciate your honesty.\end{minipage}

  \noindent\hspace*{1.4em}\begin{minipage}{\dimexpr\linewidth-1.4em\relax}4. \textbf{Use Time Blocks}: Consider using techniques like the Pomodoro Technique, where you work for a set period (like 25 minutes) and then take a short break (5 minutes). During those breaks, you could have a quick chat without feeling guilty about interrupting your workflow.\end{minipage}

  \noindent\hspace*{1.4em}\begin{minipage}{\dimexpr\linewidth-1.4em\relax}5. \textbf{Mindful Engagement}: When you do engage in conversation, try to be fully present, but also keep the conversations concise. You can have meaningful talks without them becoming too lengthy.\end{minipage}

  \noindent\hspace*{1.4em}\begin{minipage}{\dimexpr\linewidth-1.4em\relax}6. \textbf{Find Balance in Shared Interests}: If possible, steer conversations toward topics that may inspire your creative projects. This way, you can enjoy the dialogue while still keeping your mind engaged in your work.\end{minipage}

  \noindent\hspace*{1.4em}\begin{minipage}{\dimexpr\linewidth-1.4em\relax}7. \textbf{Reflect on Priorities}: Assess the urgency and importance of your projects. If a deadline is approaching, it may be necessary to prioritize your work over conversations temporarily.\end{minipage}

  \noindent\hspace*{1.4em}\begin{minipage}{\dimexpr\linewidth-1.4em\relax}By finding a balance that respects both your need for connection and your need for focus, you can cultivate both meaningful relationships and a productive work environment.\end{minipage}

\end{samplecard}

\end{document}